\documentclass[preprint,12pt]{elsarticle}

\usepackage{amssymb}
\usepackage{float}
\usepackage{subcaption}
\usepackage{hyperref}
\usepackage{amsmath}
\usepackage{tikz}

\usetikzlibrary{positioning,arrows.meta}
\usepackage{graphicx} 
\usepackage{booktabs}
\usepackage{multirow}

\usepackage[ruled,vlined]{algorithm2e}

\journal{Expert Systems With Applications}

\begin{document}

\begin{frontmatter}


\title{Deep learning based Non-Rigid Volume-to-Surface Registration for Brain Shift compensation Using Point Cloud}

\author[acmit,muv]{Eashrat Jahan Muniya} 
\ead{eashrat.muniya@acmit.at}
\author[acmit]{Gernot Kronreif}
\ead{gernot.kronreif@acmit.at}
\author[DAMTP]{Ander Biguri}
\ead{ab2860@cam.ac.uk}
\author[muv]{Wolfgang Birkfellner}
\ead{wolfgang.birkfellner@meduniwien.ac.at}
\author[dpu,acmit,muv]{Sepideh Hatamikia\corref{cor1}}
\ead{Sepideh.Hatamikia@dp-uni.ac.at}
\cortext[cor1]{Corresponding author}

\affiliation[acmit]{organization={Austrian Center for Medical Innovation and Technology},
            city={Wiener Neustadt},
            country={Austria}}
\affiliation[muv]{organization={ Medical University of Vienna},
Department and Organization= {Department of Medical Physics and Biomedical Engineering},
            city={Vienna},
            country={Austria}}
\affiliation[DAMTP]{organization={University of Cambridge}, 
Department and Organization= {Department of Applied Mathematics and Theoretical Physics (DAMTP)},
city={Cambridge}, 
country ={United Kingdom}}           

\affiliation[dpu]{organization={Danube Private University (DPU)},
Department and Organization= { Department of Medicine},
            city={Krems},
            country={Austria}}

\begin{abstract}

Soft-tissue deformation remains a major limitation of image-guided neurosurgery, where intra-operative anatomy can deviate substantially from pre-operative imaging due to brain shift, compromising navigation accuracy and surgical safety. Existing compensation methods often rely on intra-operative MRI, CT or ultrasound, which are disruptive and difficult to integrate repeatedly into the surgical workflow. In neurosurgical procedures, a partial 3D surface can be reconstructed as a point cloud from information captured by stereoscopic microscopes or laser range scanners (LRS) acquired from different viewing angles, which are derived from the visual access to a limited portion of the exposed cortex. This makes point cloud registration a meaningful approach in this setting without interrupting surgery. However, such partial and noisy surface observations make deformation estimation particularly challenging.

In this study, we propose a novel deep learning–based framework for non-rigid volume-to-surface registration, enabling the estimation of dense displacement fields from sparse intra-operative surface observations without explicit point correspondences or volumetric intra-operative imaging. The proposed network leverages multi-scale point-based feature extraction and a hierarchical deformation decoder to model both global and local deformations; however, the key contribution lies in the integration of partial intra-operative surface information into the full pre-operative point cloud domain, allowing implicit correspondence learning and dense deformation recovery under limited visibility. Quantitative results using Endpoint Error (EPE) of $1.13 \pm 0.75$  mm and RMSE of $1.33 \pm 0.81$ mm demonstrate stable performance and accurate recovery of fine-scale deformations under challenging partial-surface conditions compared with state-of-the-art non-rigid volume to surface point cloud registration methods. The proposed approach supports automatic, workflow-compatible brain-shift compensation from sparse surface observations.


\end{abstract}





\end{frontmatter}



\section{Introduction}
\label{sec1}
Image-guided surgery relies on accurate spatial correspondence between pre-operative imaging and the intra-operative anatomy. In neurosurgical procedures in particular, pre-operative magnetic resonance imaging (MRI), computed tomography (CT), and functional imaging provide high-resolution structural information that supports planning, targeting, and risk assessment \cite{liao2010automatic,hastreiter2004strategies,pereira2016volumetric}. However, once the cranium is opened, the brain undergoes complex non-rigid deformations caused by gravity, cerebrospinal fluid drainage, swelling, resection, and patient-specific biomechanical factors. This phenomenon, commonly referred to as brain shift \cite{kelly1986computer,schulz2012intraoperative}, which is also known as brain deformation, arises from a combination of biological, physical, and surgical factors and affects both cortical and deep brain structures \cite{gerard2017brain, mitsui2011skin, hill1998measurement,hammoud1996use}. To address brain shift compensation, existing methods aim to update pre-operative images based on intra-operative tissue deformation, often relying on imaging modalities such as MRI (iMR), ultrasound (iUS), laser range scanning (LRS), or stereo vision with pre-operative MRI \cite{yang2017stereovision,sinha2003cortical,xiao2019evaluation,clatz2005robust}.
Based on these inputs, a range of registration strategies have been developed to estimate the underlying deformation including physics-based biomechanical models, image-based registration methods, and point cloud–based techniques. Several studies have demonstrated that biomechanical models can provide physically plausible deformation fields and meaningful interpretability. However, their reliance on such physics based biomechanical models on patient-specific material parameters, complex meshing, and iterative solvers often results in high computational cost. Moreover, these models typically require intra-operative volumetric inputs or sparse displacement measurements, limiting their robustness when only partial surface information is available . 

A variety image-based non-rigid registration has been extensively studied using both traditional intensity-based methods and learning-based voxel-level registration frameworks. Intensity-based image registration methods attempt to directly align pre-operative and intra-operative images using intensity- or feature-based similarity measures  attempt to directly align pre-operative and intra-operative images using intensity- or feature-based similarity measures \cite{xiao2019evaluation,clatz2005robust,leroy2020intraoperative}, 
Despite their effectiveness, intensity-based methods rely on the availability of dense intra-operative imaging with sufficient contrast and spatial coverage. Furthermore, intensity-based optimisation can be sensitive to noise, modality differences, and missing structures, particularly when large deformations or resection cavities are present \cite{bayer2017intraoperative}.

Deep learning has significantly advanced deformable image registration through voxel-based deep learning networks that directly predict dense displacement vector fields from image pairs. Models such as VoxelMorph \cite{balakrishnan2019voxelmorph}, learn deformation fields in a supervised or unsupervised manner, enabling fast inference and scalable training across large datasets  \cite{fu2020deep}.
While voxel-based deep learning approaches achieve impressive performance in image-to-image registration, they inherit the fundamental assumption of dense volumetric intra-operative data. This assumption limits their applicability in neurosurgical workflows where intra-operative imaging is sparse or unavailable. Moreover, voxel-based representations are not naturally aligned with surface-based measurements acquired during surgery, making it difficult to directly exploit partial cortical surface information \cite{dalca2018unsupervised,de2019deep}. 
All the mentioned techniques above are dependent on volumetric images taken during the intervention.
\begin{figure}[H]
  \centering
  \includegraphics[width=0.90\linewidth]{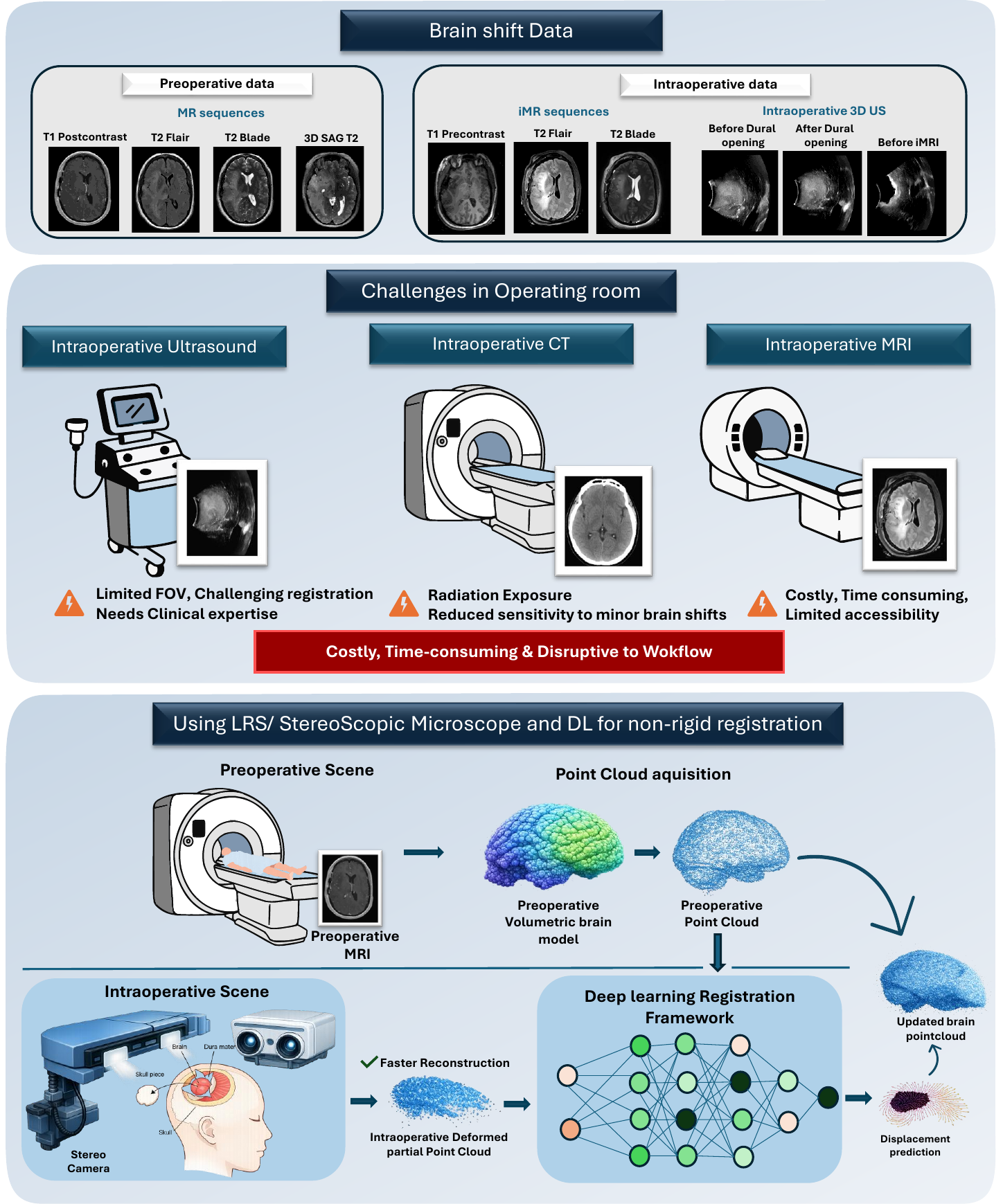}
  \caption{Schematic overview of brain-shift data, limitations of intra-operative imaging modalities, and the proposed learning-based volume-to-surface registration framework integrating pre-operative MRI, intra-operative surface acquisition, and deep neural network-based deformation field estimation.}
  \label{schematic}
\end{figure}

Intra-operative imaging modalities such as magnetic resonance imaging (MRI) and ultrasound (US) enable visualization of brain deformation during surgery and have been shown to improve tumor resection when combined with advanced registration techniques \cite{clatz2005robust,xiao2019evaluation}. However, intra-operative MRI is expensive, available in only a limited number of operating rooms, and the acquisition of advanced sequences can disrupt surgical workflow \cite{correa2017neurosurgery}. Ultrasound imaging can be performed intra-operatively with minimal time overhead, but its limited image quality, restricted field of view, and differences from pre-operative MRI make accurate registration challenging and highly operator-dependent \cite{correa2017neurosurgery,yang2017stereovision,zhang2019wireless}. These limitations have motivated alternative approaches based on partial surface data acquired intra-operatively using laser range scanners (LRS) or stereoscopic microscopes, \cite{yang2017stereovision,sinha2003cortical}. These systems can be integrated into the operating room and provide high-resolution three-dimensional point clouds of the exposed cortical surface with minimal workflow disruption. Such partial surface observations can be registered to pre-operative models to estimate brain shift and deformation, highlighting the need for robust non-rigid point cloud registration methods capable of handling sparse and incomplete data. An overview of the clinical workflow, the limitations of intra-operative imaging, and the proposed learning-based volume-to-surface registration
concept is illustrated in Fig.~\ref{schematic}.

Surface-based registration has recently gained attention as an alternative to volumetric alignment in scenarios where intra-operative imaging is limited \cite{yang2025resolving,liu2024non,jiang1992new, labrunie2022automatic, liu2023ranerf, zhang2024point}. Early approaches primarily rely on variants of the Iterative Closest Point (ICP)\cite{besl1992method} algorithm, which estimate rigid or non-rigid NICP \cite{amberg2007optimal}transformations by minimizing point-to-point or point-to-surface distances. Although computationally efficient, ICP-based methods are highly sensitive to initialization, noise, and outliers, and assume substantial surface overlap and reliable correspondences. These assumptions are rarely satisfied in neurosurgical settings, where only partial cortical surfaces are visible. Recent learning-based methods aim to infer volumetric deformation from sparse surface observations in the absence of dense intra-operative imaging. By conditioning deformation models on partial surface information, these approaches provide data-driven alternatives to classical interpolation and biomechanical simulation. PointNet \cite{qi2017pointnet, qi2017pointnetplusplus} Dynamic Graph CNN (DGCNN) \cite{phan2018dgcnn}, DeepGCNs \cite{li2019deepgcns} architectures enable direct feature learning from unordered point sets and have shown robustness to moderate deformation and noise. A comprehensive overview of deep learning approaches for point cloud registration and their taxonomy is provided in recent survey literature \cite{zhang2024comprehensive}. These architectures enable direct feature learning from unordered point sets and have shown robustness to moderate deformation and noise.  However, most existing frameworks focus on surface-to-surface alignment and typically estimate global transformations or sparse correspondences, limiting their ability to recover dense volumetric deformation from partial observations \cite{miga2000model,aoki2019pointnetlk, huang2021predator} which is essential for accurate brain-shift compensation under limited visibility.

In non-neurosurgical applications, learning-based volume-to-surface registration has shown promising results. For organs such as the liver \cite{henrich2025ludo}, lungs \cite{falta2023lung250m}, ear \cite{liu2024non}, and prostate \cite{prananta2010robotic, haker2004landmark} intra-operative surfaces or sparse landmarks often exhibit smoother, more globally correlated deformation patterns, and surface visibility is typically more extensive. Under these conditions, neural networks can learn mappings between surface motion and volumetric deformation using synthetic simulations, enabling reasonable deformation recovery within the organ interior. However, to date, such learning-based volume-to-surface approaches have not been successfully applied to brain shift compensation. Brain shift presents unique challenges due to severely limited and asymmetric surface visibility, highly heterogeneous and localized deformation patterns, and the need to infer displacement in large unobserved regions without direct intra-operative measurements. 

Many existing methods aim to align intra-operative surface observations with the pre-operative anatomy through surface-based registration, landmark matching, or correspondence estimation \cite{fan2020robust, wang2024brainmorph, 10304319, 6193441}. These approaches primarily optimise the alignment of the observed cortical surface and therefore focus on surface correspondence rather than volumetric deformation recovery. However, brain shift compensation requires estimating the displacement field throughout the brain volume, including large regions where no direct intra-operative observations are available.

Motivated by this gap, we propose a learning-based volume-to-surface non-rigid registration framework specifically designed for brain shift compensation. The method explicitly models the asymmetry between dense pre-operative anatomical models and sparse, partial intra-operative surface observations represented as point clouds. A multi-scale PointNet-based hierarchical deformation decoder is used to learn geometric priors from synthetic brain shift simulations and to predict dense displacement vector fields (DVFs) from incomplete cortical surface measurements. This design enables robust deformation estimation under limited visibility, noise, and missing correspondences, addressing key challenges that have prevented prior volume-to-surface learning approaches from being applied to brain shift. Experimental results demonstrate consistent improvements over state-of-the-art learning-based and classical registration methods across metrics including EPE and RMSE. The main contributions of this work are as follows:

\begin{itemize}

\item We introduce a learning-based volume-to-surface registration framework tailored for brain shift compensation that explicitly models the asymmetry between dense pre-operative anatomy and sparse intra-operative cortical surface observations.

\item The method operates without the need for explicit one-to-one correspondences between the pre-operative point cloud and the partial intra-operative surface observations.

\item We design a training strategy based on synthetic brain shift simulations that enables the network to learn geometric deformation and recover volumetric deformation even in large unobserved regions.

\end{itemize}


\section{Methods}
\begin{figure}[H]
  \centering
  \includegraphics[width=1.0\linewidth]{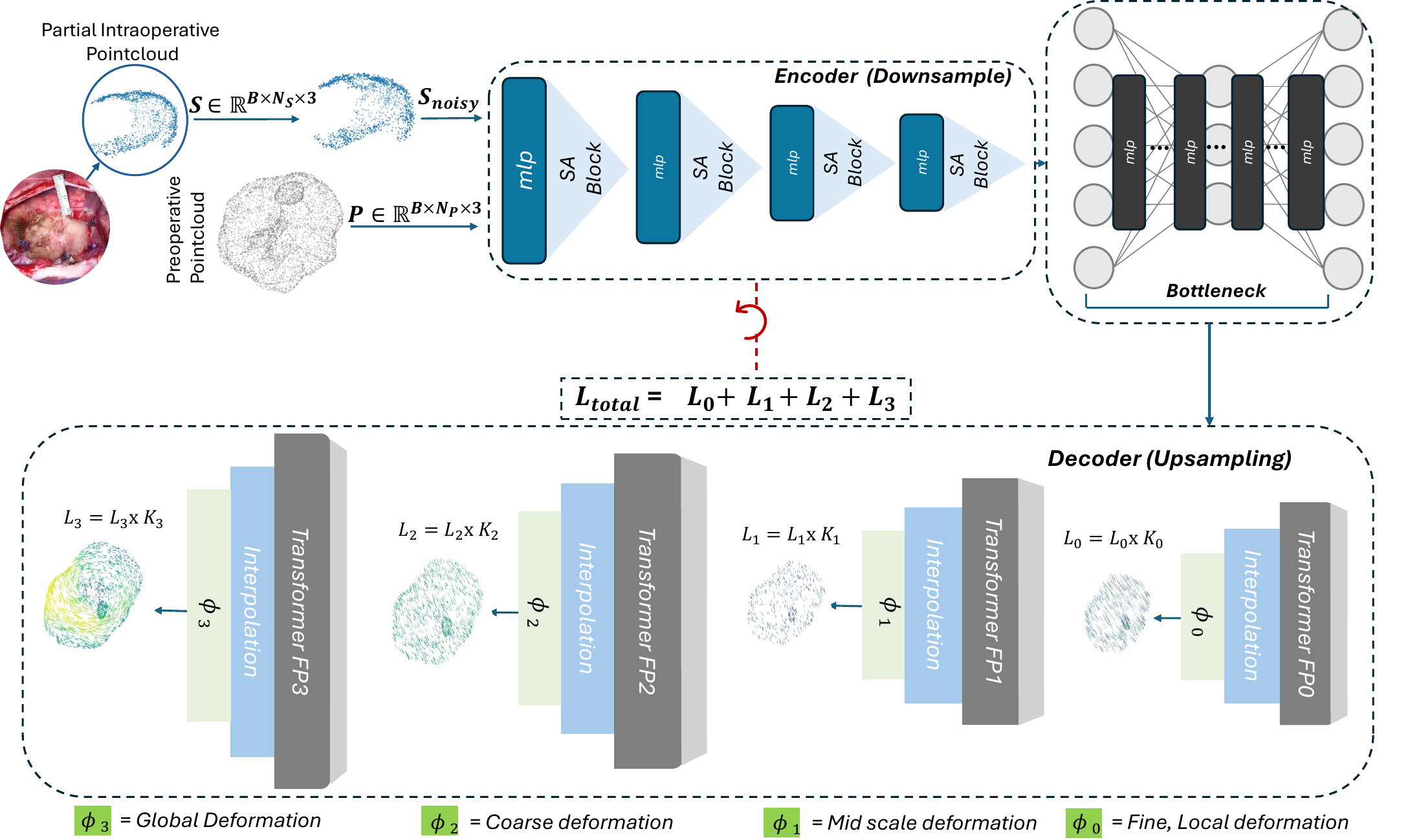}
  \caption{Overview of the proposed volume-to-surface registration network. The network follows an encoder--decoder architecture, where the encoder extracts hierarchical features from the preoperative point cloud $P$ and the partial intraoperative surface $S$ using successive set abstraction (SA) blocks, followed by a bottleneck. The decoder progressively upsamples features via interpolation and Transformer-based feature propagation modules (FP0--FP3) to predict multi-scale deformation fields $\phi_0$--$\phi_3$, ranging from local to global deformations. The total loss is defined as the weighted sum across all scales.}
  \label{architechture}
\end{figure}

Let us assume, 

\begin{equation}
\mathcal{P} = \{ \mathbf{p}_i \in \mathbb{R}^3 \}_{i=1}^{N}
\label{eq:preop_points}
\end{equation}

denotes the pre-operative point cloud, and

\begin{equation}
\mathcal{S} = \{ \mathbf{s}_j \in \mathbb{R}^3 \}_{j=1}^{M}
\label{eq:intraop_points}
\end{equation}

where $\mathcal{P}$ denotes the pre-operative volumetric point cloud and $\mathcal{S}$ represents a partial intra-operative cortical surface observation. The two point sets are sampled independently from different anatomical states and therefore do not share a one-to-one correspondence. In particular, $\mathcal{S}$ is not a subset of $\mathcal{P}$ but rather a sparse observation of the deformed cortical surface obtained intra-operatively can be denoted as $\mathcal{I}$. The objective of the proposed method is to estimate a volumetric displacement field that aligns the pre-operative anatomy with the observed intra-operative surface without requiring explicit point correspondences.
The goal is to estimate a deformation field: 

\begin{equation}
\phi : \mathbb{R}^3 \rightarrow \mathbb{R}^3
\end{equation}
which maps points from the preoperative configuration to their intraoperative locations, while remaining consistent with the observed partial surface geometry and enforcing physically plausible deformation behavior.
We propose a learning-based framework for non-rigid volume-to-surface registration, where a complete preoperative point cloud $P \in \mathbb{R}^{B \times N_P \times 3}$ is aligned to a partial intraoperative surface observation $S \in \mathbb{R}^{B \times N_S \times 3}$. No explicit point-wise correspondences between $P$ and $S$ are assumed. The overall architecture is illustrated in figure \ref{architechture}.


\begin{figure}[H]
  \centering
  \includegraphics[width=1.0\linewidth]{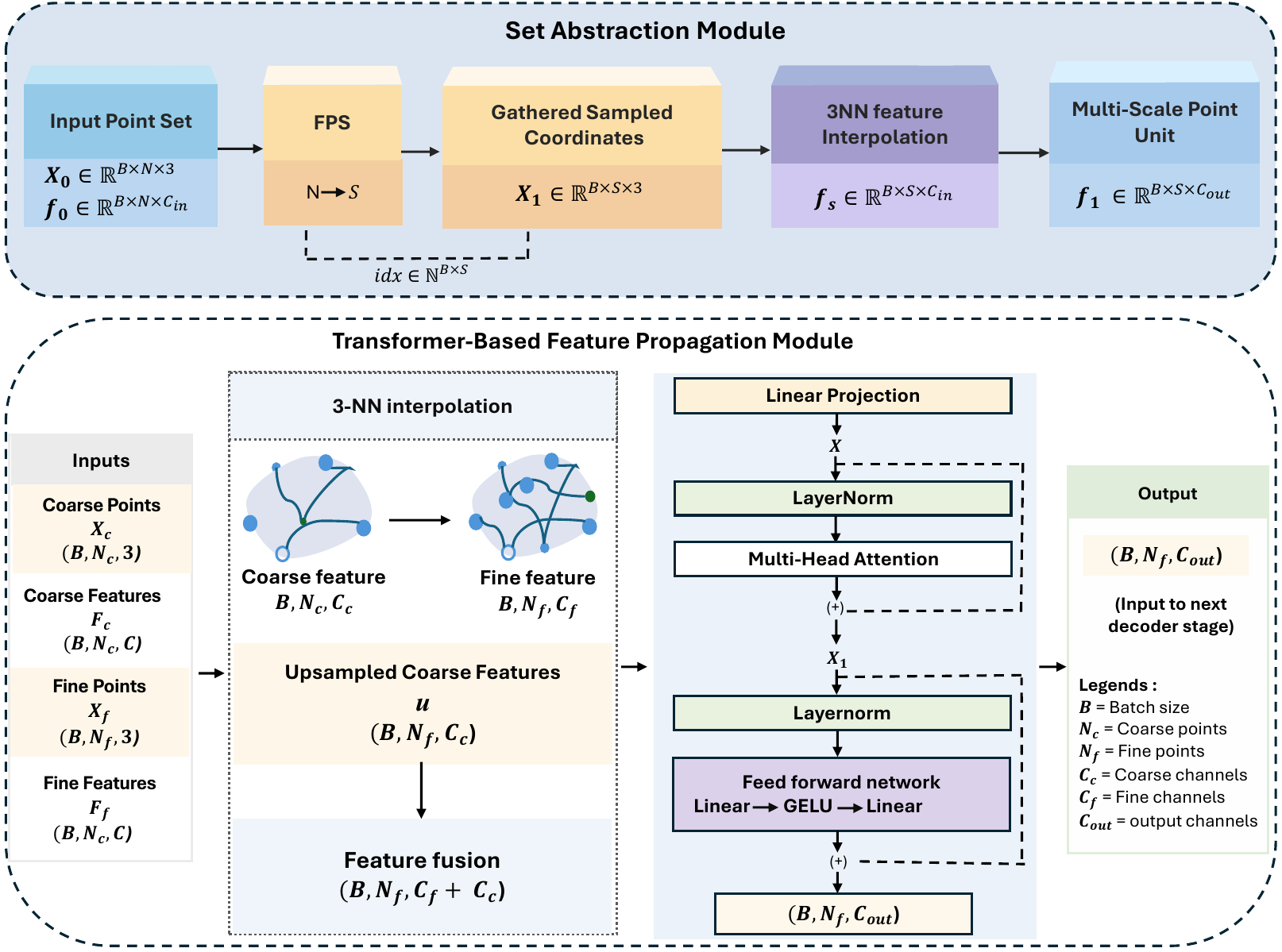}
  \caption{Set abstraction (top) and Transformer-based feature propagation (bottom) modules. The SA module performs hierarchical downsampling using farthest point sampling (FPS), followed by feature interpolation and local aggregation via a multi-scale point unit to produce coarse representations. The FP-Transformer module propagates features from coarse to fine resolutions using 3-NN interpolation, fuses them with fine-level features, and refines the representation using a Transformer encoder with residual connections. The output corresponds to refined fine-level features used in subsequent decoding stages.}
  \label{architechture2}
\end{figure}


\textbf{Architecture overview:}
The network follows an encoder--decoder design that learns hierarchical geometric features and predicts deformation fields in a coarse-to-fine manner. Such a design is widely adopted in deformable registration, as it enables the model to first capture large-scale structural variations at coarse resolutions and subsequently refine finer geometric details at higher resolutions. The encoder encodes multi-scale geometric context, while the decoder progressively refines deformation estimates at increasing spatial resolutions.

\textbf{Encoder: hierarchical feature extraction:}
The encoder processes the input point clouds using a sequence of Set Abstraction (SA) modules \cite{qi2017pointnetplusplus} (Fig.~3, top). Given input coordinates and features $(x_0, f_0)$, where $x_0 \in \mathbb{R}^{B \times N \times 3}$ and $f_0 \in \mathbb{R}^{B \times N \times C_{in}}$, each SA module performs farthest point sampling (FPS) to obtain a reduced set of points $x_1 \in \mathbb{R}^{B \times S \times 3}$. Features are transferred to the sampled points via 3-nearest neighbor interpolation, yielding $f_s \in \mathbb{R}^{B \times S \times C_{in}}$, followed by local geometric aggregation using a multi-scale point unit to produce refined features $f_1 \in \mathbb{R}^{B \times S \times C_{out}}$. Stacking multiple SA levels constructs a hierarchical representation $\{(x_l, f_l)\}_{l=0}^{L}$, where spatial resolution decreases while feature abstraction increases. This allows the network to capture both local surface structure and global anatomical context.



\textbf{Multi-scale feature learning}
Multi-scale feature learning is incorporated within each Set Abstraction (SA) module through the Multi-Scale Point Unit, as shown in Fig.~3 (top). This unit performs local geometric aggregation over neighborhoods of different spatial extents, enabling the encoder to capture both fine-scale surface details and broader structural context. Such multi-scale aggregation is particularly beneficial when only partial intraoperative observations are available.


\textbf{Bottleneck representation:}
At the coarsest level, the encoder produces a compact latent representation summarizing the global geometry of the preoperative anatomy and its relationship to the observed intraoperative surface. This representation provides global context for subsequent deformation estimation.

\textbf{Decoder transformer-based feature propagation:}
The decoder reconstructs point-wise features at higher resolutions using a sequence of feature propagation blocks (Fig.~3, bottom). At each stage, coarse features $(x_c, f_c)$ are interpolated onto a finer point set $x_f$ using 3-NN interpolation:
\begin{equation}
\mathbf{u} \in \mathbb{R}^{B \times N_f \times C_c}
\end{equation}
The interpolated features are concatenated with fine-level features $f_f$ to form fused representations in
\begin{equation}
\mathbb{R}^{B \times N_f \times (C_f + C_c)}
\end{equation}

Unlike conventional PointNet++ decoders that rely on shared MLPs, we employ a Transformer encoder to refine these fused features. The Transformer models long-range dependencies across points via self-attention, enabling global reasoning about deformation patterns. Residual connections and feed-forward layers further stabilize feature refinement.

\textbf{Multiscale deformation prediction:}
Deformation fields are predicted at multiple decoder stages. At each level $k$, the network outputs a displacement field
\begin{equation}
u_k(x) \in \mathbb{R}^3
\end{equation}
representing point-wise 3D displacements at that resolution. Early decoder stages capture coarse, global deformations, while later stages refine local, high-frequency motion.

\textbf{Output deformation field and training objective:}
The final deformation field is obtained from the highest-resolution prediction:
\begin{equation}
\phi(x) = u_K(x)
\end{equation}
which provides a dense displacement field aligning the preoperative point cloud to the intraoperative observation. The network is trained in a fully supervised manner, where the predicted displacement fields are directly compared with the ground-truth deformation field. To stabilize training and enforce consistency across multiple resolutions, deep supervision is applied \cite{pfeiffer2019learning}. The total loss is defined as:
\begin{equation}
\mathcal{L}_{\text{total}} = \sum_{k=0}^{K} \lambda_k \, \mathcal{L}_k
\end{equation}
where $\mathcal{L}_k$ denotes the loss at scale $k$, computed between the predicted displacement field $u_k(x)$ and the corresponding ground-truth deformation field $\phi_{\text{gt}}(x)$, and $\lambda_k$ are weighting coefficients.

Each scale-specific loss is defined as:
\begin{equation}
\mathcal{L}_k = \frac{1}{N} \sum_{i=1}^{N} \left\| u_k(x_i) - \phi_{\text{gt}}(x_i) \right\|_2
\end{equation}

The final predicted deformation field $\phi(x)$ is then used to warp the preoperative point cloud, enabling alignment with the observed intraoperative surface.
\subsection{Dataset}
\begin{figure}[H]
  \centering
  \includegraphics[width=\linewidth]{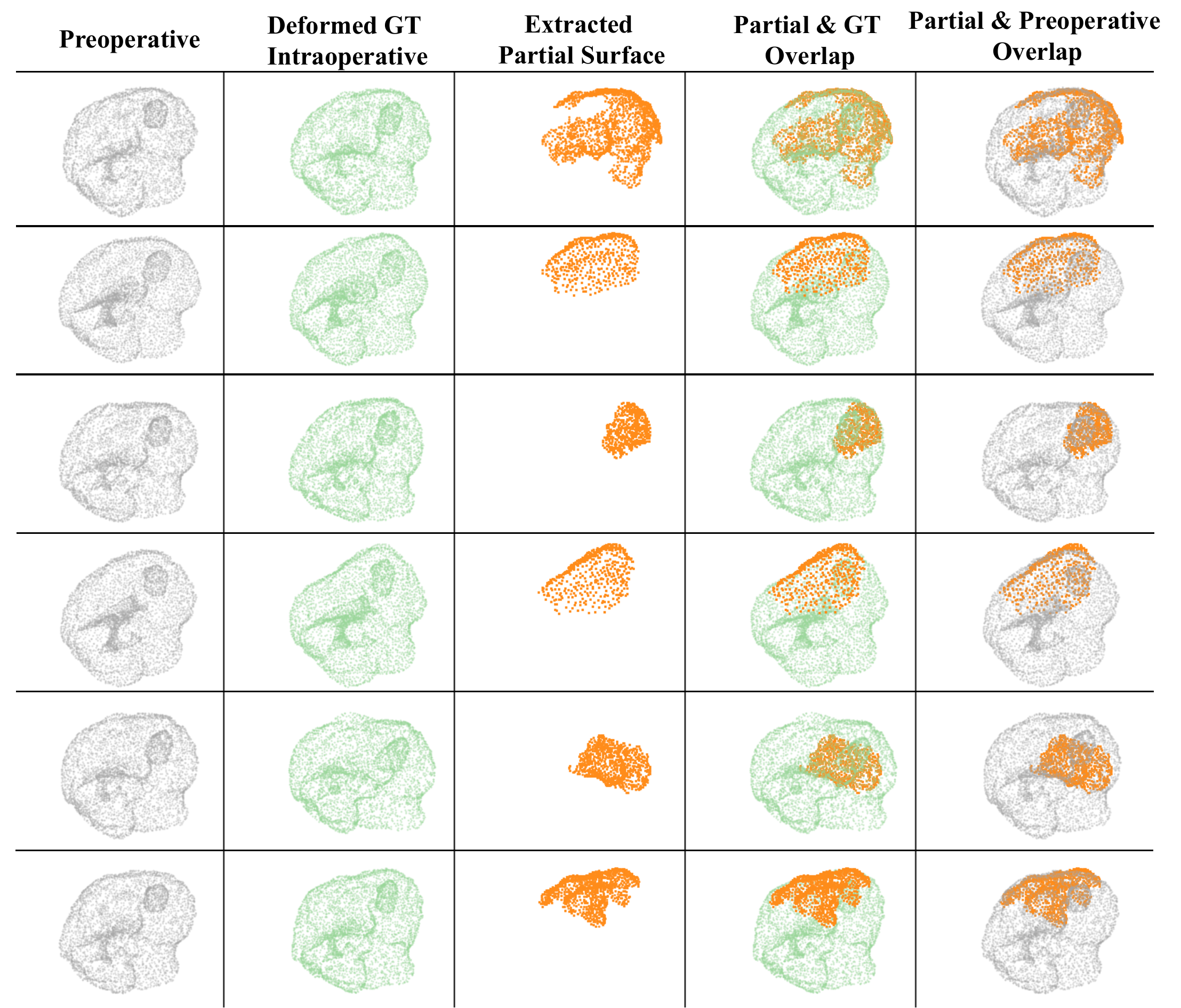}
  \caption{Visualization of the training data generation for the proposed volume-to-surface registration framework across multiple cases. For each example, the pre-operative point cloud ($P$, gray) is synthetically deformed to produce the ground-truth intra-operative configuration ($I_{gt}$, green). A partial surface ($S$, orange) is extracted from $I_{gt}$ to simulate intra-operative surface observations. The final two columns show the overlap between $S$ and $I_{gt}$, and between $S$ and $P$, respectively, illustrating the input configuration used for training.}
  \label{fig:data}
\end{figure}

A publicly available brain MRI dataset from The Cancer Imaging Archive (TCIA), specifically the Brain Resection Multimodal Imaging Database (ReMIND) \cite{juvekar2024remind}, was used to obtain patient-specific preoperative anatomical models. In this work, only the preoperative imaging data were utilized, while intraoperative observations and deformation fields were synthetically generated.
For each case, the preoperative MRI was processed to extract a brain surface mesh, which was subsequently represented as a dense point cloud. Synthetic deformations and partial intraoperative observations were generated for training, illustrated in Fig.~\ref{fig:data}. Further details on data generation and deformation modeling are provided in Section \ref{datagen}.

\subsubsection{Data Representation}

In this study, the preoperative point cloud represents the anatomical configuration of the patient’s brain prior to surgery, including the cortical surface, tumor, and ventricular structures. This point cloud serves as the undeformed reference state.

The intraoperative point cloud corresponds to a deformed configuration of the brain surface under different deformation constraints. To reflect the limited visibility typically available during neurosurgical procedures, only a subset of the deformed cortical surface is extracted and treated as a partial point cloud. This setting mimics the intraoperative scenario, where only a restricted portion of the cortex is observable.
The objective of registering the preoperative point cloud $\mathcal{P}$ with the partial intraoperative surface $\mathcal{S}$ is to estimate a dense volumetric deformation field $\phi(\mathbf{x})$. Specifically, we aim to infer how the preoperative anatomy deforms intraoperatively under given deformation constraints, despite having access to only sparse and partial surface observations.

\subsection{Data Generation}
\label{datagen}
To enable supervised learning and controlled evaluation of volume-to-surface registration under partial observability, we construct a synthetic dataset of preoperative and intraoperative brain configurations with known ground-truth deformations.
A total of 10 patients were used, from which approximately 1500 synthetic deformations per case were generated, yielding around 15{,}000 samples. The dataset was split into 70\% training, 15\% validation, and 15\% testing sets. Each preoperative model was represented by approximately $N_p$ points. For each sample, a dense intraoperative point cloud $I \in \mathbb{R}^{N_p \times 3}$ was generated via synthetic deformation, maintaining a one-to-one correspondence with $P$. Partial intraoperative observations $S \in \mathbb{R}^{N_s \times 3}$ were then extracted from $I$ at different visibility levels (vis25--vis65), with $N_s < N_p$. Each sample thus consists of $(P, I, S, \Phi)$, where $P \in \mathbb{R}^{N_p \times 3}$ denotes the preoperative point cloud, $I \in \mathbb{R}^{N_p \times 3}$ the dense intraoperative configuration, $S \in \mathbb{R}^{N_s \times 3}$ the partial intraoperative surface, and $\Phi \in \mathbb{R}^{N_p \times 3}$ the corresponding displacement field. Details of the synthetic data generation pipeline are provided in the following subsections.

\subsubsection{Anatomical Model Construction} 
\begin{figure}[H]
  \centering
  \includegraphics[width=0.65\linewidth]{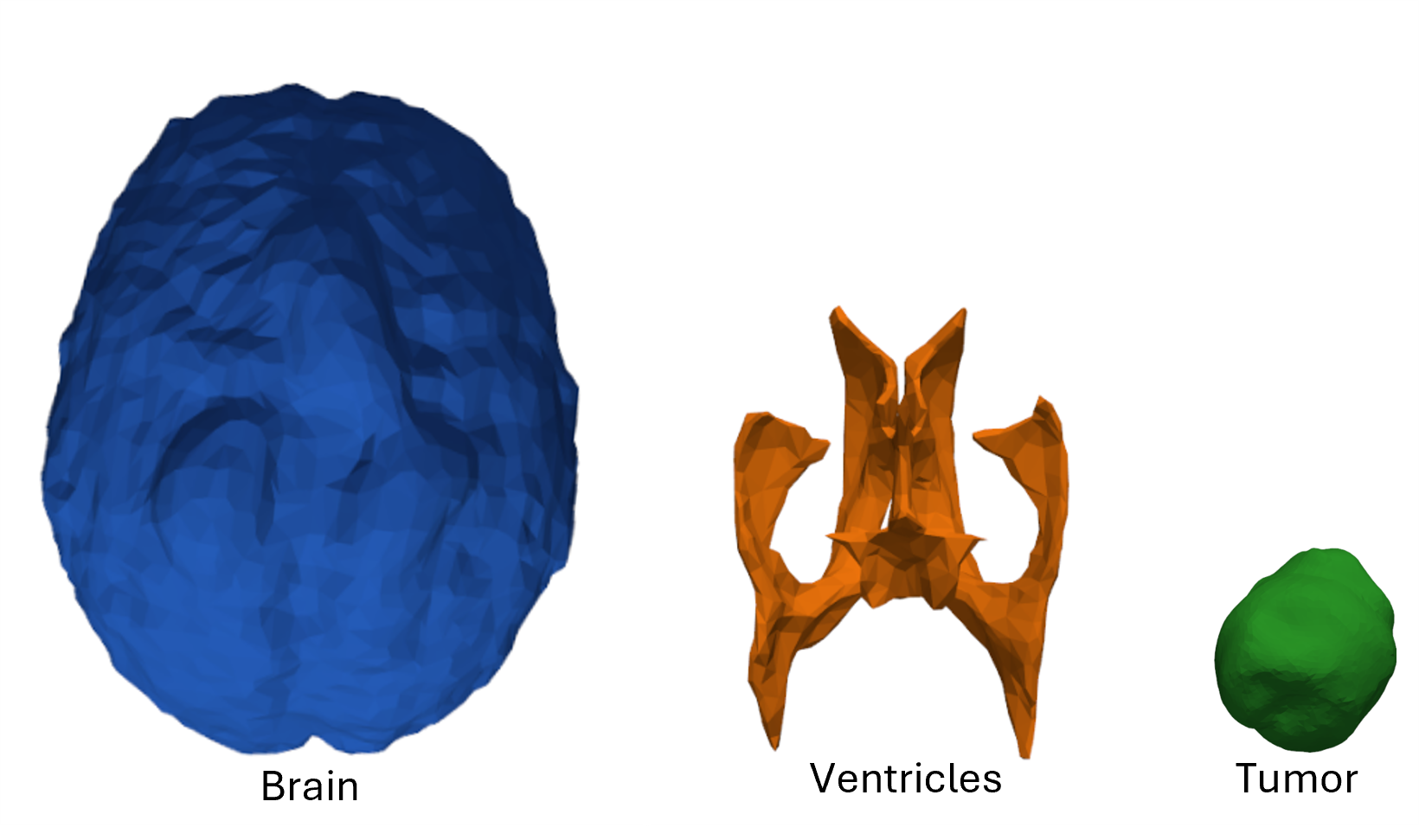}
  \caption{Simplified surface mesh models of the anatomical components used in this study, including the brain surface (blue), ventricular system (orange), and tumor region (green).}
  \label{figbrainanatomy}
\end{figure}

\begin{figure}[H]
  \centering
  \includegraphics[width=0.65\linewidth]{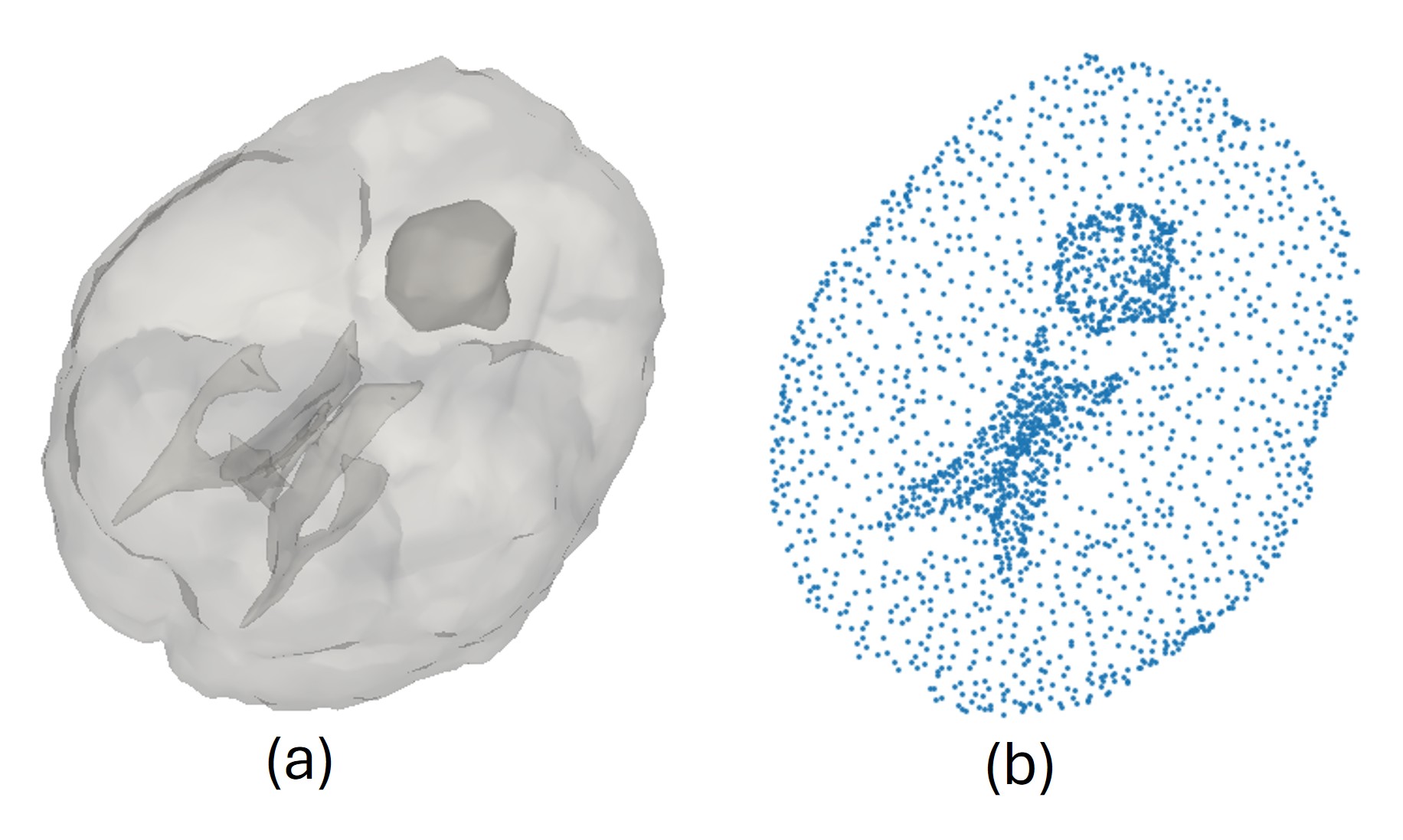}
  \caption{(a) Visualization of the combined anatomical model used for dataset generation. The tumor and ventricular structures are integrated within the brain surface to form a unified geometric representation . (b) Illustrates the corresponding sampled point cloud representation employed for learning-based deformation modeling.}
  \label{combinedfig}
\end{figure}

A high-resolution preoperative brain MRI from the REMIND dataset is first segmented using 3D Slicer \cite{3d}, to obtain separate volumetric segmentations of the brain, tumor, and ventricles. Each segmentation is converted into an individual 3D surface mesh with Gmesh \cite{geuzaine2009gmsh}  and meshmixer \cite{meshmixer} which is subsequently represented as a point cloud. Individual meshes of the brain surface, ventricular, and tumor region were preprocessed can be seen in figure \ref{figbrainanatomy} and spatially aligned before being integrated into a unified composite model figure \ref{combinedfig} (a).
The segmented brain, tumor, and ventricle models are then digitally injected, spatially aligned and merged into a single unified anatomical representation. This composite model serves as the preoperative reference configuration and captures both cortical and internal anatomical structures relevant to brain-shift modeling.

\subsubsection{Deformation Simulation} To generate intraoperative configurations, synthetic non-rigid deformations are applied to the unified preoperative model. These deformations are designed to mimic plausible intraoperative brain shift patterns while preserving anatomical coherence across cortical and subsurface structures. The corresponding pseudocode is presented in algorithm \ref{alg:deformation_generation_simple}.

\textbf{Analytic deformation operators:}
We generate synthetic deformations using analytically defined operators applied directly to mesh vertices.
Let $\mathbf{v}\in\mathbb{R}^3$ denote a vertex, and let $\mathbf{n}(\mathbf{v})$ be the corresponding unit vertex normal.

\textbf{(i) Bulge operator $\mathcal{B}$ (normal-directed Gaussian bump)}
A localized bulge centered at a point $\mathbf{c}_{\text{b}} \in \mathbb{R}^3$ is defined as
\begin{equation}
\mathcal{B}(\mathbf{v}) = \mathbf{v} + s\,\exp\!\left(-\left(\tfrac{\|\mathbf{v}-\mathbf{c}_{b}\|}{r}\right)^2\right)\,\mathbf{n}(\mathbf{v})
\end{equation}
where $r$ controls the spatial extent and $s$ controls the deformation magnitude.

\textbf{(ii) Sliding operator $\mathcal{S}$ (plane-gated translation)}
Given a plane defined by a point $\mathbf{p}$ and unit normal $\hat{\mathbf{n}}$, we compute signed distance
$d(\mathbf{v}) = (\mathbf{v}-\mathbf{p})^\top \hat{\mathbf{n}}$ and apply a smooth shift
\begin{equation}
\mathcal{S}(\mathbf{v}) = \mathbf{v} + \exp\!\left(-\left(\tfrac{\max(0,d(\mathbf{v}))}{w}\right)^2\right)\,\Delta\mathbf{t}
\end{equation}
where $w$ controls the transition width and $\Delta\mathbf{t} \in \mathbb{R}^3$ is a displacement vector defining the direction and magnitude of the translation. The deformation is applied only to vertices with positive signed distance to the plane and is smoothly attenuated with increasing distance from the plane.

\textbf{(iii) Twist operator $\mathcal{T}$ (radius-decayed rotation around an axis)}
Let the twist axis be defined by a point $\mathbf{a} \in \mathbb{R}^3$ and a unit direction $\mathbf{u} \in \mathbb{R}^3$.
For each vertex, we compute the component of $(\mathbf{v}-\mathbf{a})$ orthogonal to the axis,
\[
\mathbf{r}_{\perp} = (\mathbf{v}-\mathbf{a}) - \big((\mathbf{v}-\mathbf{a})^\top \mathbf{u}\big)\mathbf{u},
\]
and its magnitude $r = \|\mathbf{r}_{\perp}\|$.
The rotation angle is then defined as
\begin{equation}
\theta(\mathbf{v})=\theta_{\max}\exp\!\left(-\left(\tfrac{r}{r_0}\right)^2\right)
\end{equation}
where $r_0$ controls the spatial decay of the twist.
The orthogonal component $\mathbf{r}_{\perp}$ is then rotated about the axis defined by $\mathbf{u}$ using Rodrigues' rotation formula.




\textbf{(iv) Global warp operator $\mathcal{W}$ (sinusoidal displacement field).}
A smooth global displacement field is defined as follows. 
For a vertex $\mathbf{v}=(x,y,z)^\top$,
\begin{subequations}\label{eq:sin_deformation}
\begin{equation}
\Delta x = a\sin(f_x\,y+\varphi_x)
\end{equation}

\begin{equation}
\Delta y = a\sin(f_y\,z+\varphi_y)
\end{equation}

\begin{equation}
\Delta z = a\sin(f_z\,x+\varphi_z)
\end{equation}
\end{subequations}

where $a$ denotes the displacement amplitude, $f_x, f_y, f_z$ are spatial frequencies, and $\varphi_x, \varphi_y, \varphi_z$ are phase offsets that shift the sinusoidal deformation along each axis.

The warped vertex is given by
\[
\mathcal{W}(\mathbf{v}) = \mathbf{v} + (\Delta x, \Delta y, \Delta z)^\top
\]
\textbf{Composite deformation:}
The composite mode applies an ordered composition of operators:
\begin{equation}
\mathcal{C}(\mathbf{v}) = \mathcal{W}(\mathcal{T}(\mathcal{S}(\mathcal{B}(\mathbf{v}))))
\end{equation}
with parameters sampled from predefined ranges and anchored using the preoperative axis-aligned bounding box (AABB) center to ensure consistent deformation across anatomical structures.

\textbf{Post-deformation smoothing:}
To reduce high-frequency artifacts introduced by the analytic deformation operators, the deformed meshes are further regularized using Taubin smoothing~\cite{taubin1995curve}. This smoothing is applied iteratively with fixed parameters, preserving the overall deformation while improving surface regularity.

\begin{algorithm}[H]
\caption{Synthetic deformation generation (Open3D-based analytic operators)}
\label{alg:deformation_generation_simple}
\DontPrintSemicolon
\KwIn{Set of mesh pairs $\{(M^{pre}_c, M^{intra}_c)\}_{c=1}^N$ ;where $c$ indexes the case and $N$ is the total number of mesh pairs ;deformation mode set $\mathcal{M}=\{\texttt{global},\texttt{combo}\}$, where \texttt{global} denotes a global non-rigid deformation, and \texttt{composite} denotes a sequential composition of deformation operators (bulge, sliding, twisting, and global warp); number of variations $V$; smoothing flag}

\KwOut{Deformed mesh pairs saved to disk}

\ForEach{mesh pair $(M^{pre}_c, M^{intra}_c)$}{
  Load meshes $M^{pre}_c$ and $M^{intra}_c$\;
  Compute vertex normals\;
  Compute AABB center $\mathbf{c}$ from $M^{pre}_c$\;

  \ForEach{mode $m \in \mathcal{M}$}{
    \For{$v \leftarrow 1$ \KwTo $V$}{
      Set deterministic seed from $(c,m,v)$ and sample parameters $\theta_{c,m,v}$\;

      \eIf{$m=\texttt{global}$}{
        $\widetilde{M}^{pre} \leftarrow \texttt{GlobalWarp}(M^{pre}_c;\theta_{c,m,v})$\;
        $\widetilde{M}^{intra} \leftarrow \texttt{GlobalWarp}(M^{intra}_c;\theta_{c,m,v})$\;
      }{
        $\widetilde{M}^{pre} \leftarrow \texttt{CompositeDeform}(M^{pre}_c;\theta_{c,m,v})$\;
       $\widetilde{M}^{intra} \leftarrow \texttt{CompositeDeform}(M^{intra}_c;\theta_{c,m,v})$\;
      }

      \If{\texttt{smoothing}}{
        $\widetilde{M}^{pre} \leftarrow \texttt{TaubinSmooth}(\widetilde{M}^{pre})$\;
        $\widetilde{M}^{intra} \leftarrow \texttt{TaubinSmooth}(\widetilde{M}^{intra})$\;
      }

      Save $(\widetilde{M}^{pre},\widetilde{M}^{intra})$ with standardized naming\;
    }
  }
}
\end{algorithm}
The applied deformation defines a dense ground-truth volumetric displacement field $\phi(\mathbf{x})$, which is used to warp the preoperative point cloud and produce a fully deformed brain model. This deformed configuration represents the intraoperative anatomy and provides exact point-wise displacement supervision for training and evaluation.

\subsubsection{Partial Surface Extraction} 
\begin{figure}[H]
  \centering
  \includegraphics[width=1.0\linewidth]{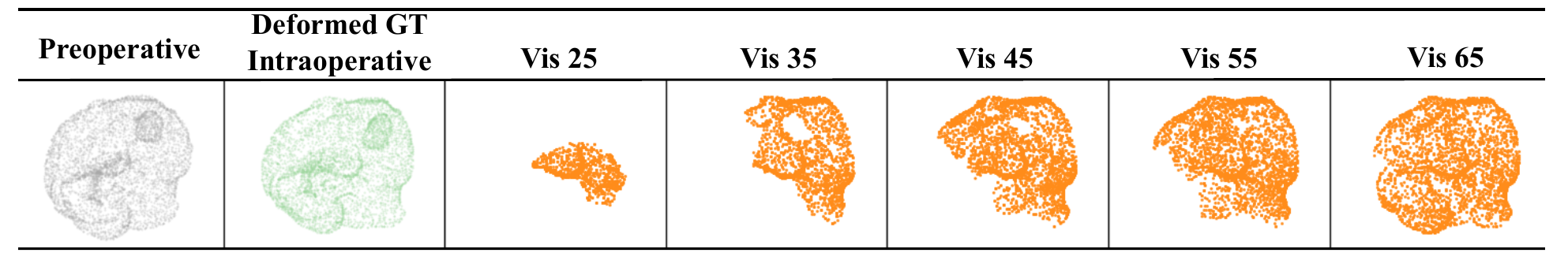}
  \caption{Visualization of partial intraoperative surface observations extracted from a single deformed brain configuration at different visibility levels. The first two panels show the preoperative point cloud and the corresponding deformed intraoperative configuration. Subsequent panels illustrate progressively extracted larger visible surface regions (vis25–vis65), simulating varying levels of cortical surface exposure during surgery.}
  \label{partial}
\end{figure}

To emulate the limited field of view available during neurosurgical procedures, only partial observations of the deformed cortical surface are retained. Specifically, random surface patches are extracted from the deformed brain surface under different visibility ratios, resulting in partial intraoperative point clouds can be seen in Figure \ref{partial}.

By varying the spatial extent and location of these extracted surface regions, multiple levels of surface visibility are simulated. This process enables systematic evaluation of registration performance under increasingly sparse and incomplete intraoperative observations, while maintaining a fixed underlying volumetric deformation.

The final dataset therefore consists of paired samples comprising:
(i) a complete preoperative point cloud,
(ii) a partial intraoperative surface point cloud(with different visibility ratio), and
(iii) the corresponding ground-truth volumetric deformation vector field.

\section{Experiments} 

\subsection{Noise Augmentation}
To simulate intra-operative surface measurement uncertainty, noise augmentation was applied only to the partial intra-operative surface point cloud $S$ during training. A composite noise model was used, consisting of (i) additive Gaussian jitter and (ii) a smooth spatially coherent displacement field, capturing both independent localization errors and correlated surface distortions arising from tissue motion and probe interaction.

\begin{figure}[H]
  \centering
  \includegraphics[width=1.0\linewidth]{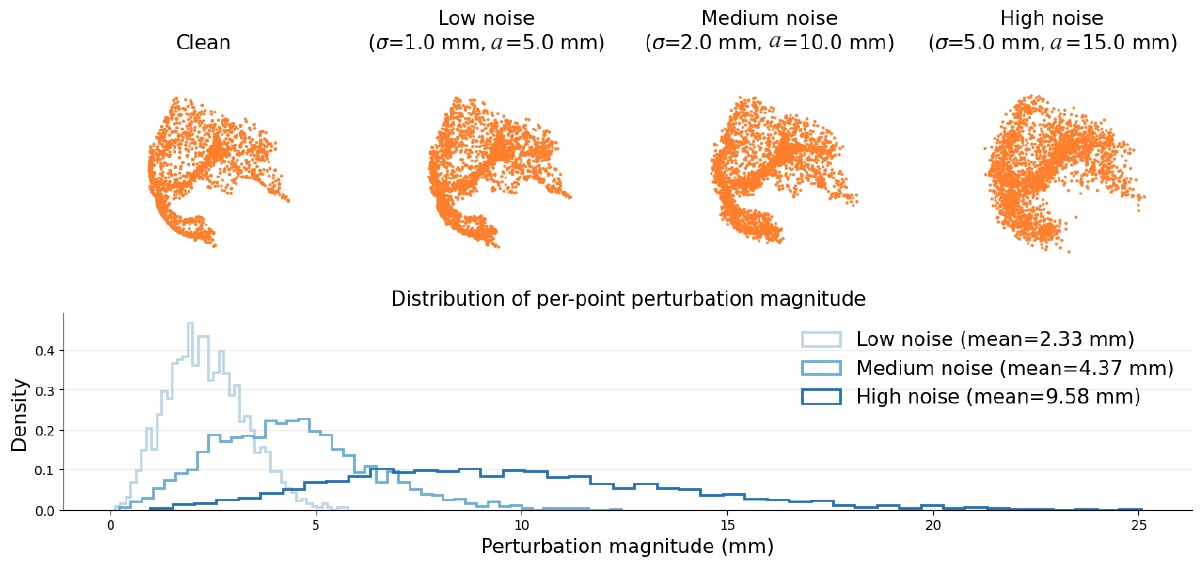}
  \caption{Surface augmentation under increasing noise levels. Top: clean and noisy point clouds. Bottom: probability density of per-point displacement magnitude (mm) between clean and noisy surfaces.}
  \label{fig:noise_levels}
\end{figure}




Let $s_i \in \mathbb{R}^3$ denote a point on the normalized surface point cloud $S$, where normalization is performed by centering at the mean of $V$ and scaling by its maximum radius. Gaussian noise is applied as
\begin{equation}
\tilde{s}_i = s_i + \epsilon_i, \quad \epsilon_i \sim \mathcal{N}(0, \sigma_n^2 I),
\label{eq:gaussian_noise}
\end{equation}
where $\sigma_n = \sigma_{mm} / (\text{scale} + \epsilon)$ and $\sigma_{mm} \sim \mathcal{U}(0, 5)$ mm.

To model spatially correlated distortions, a smooth displacement field $d(s)$ is defined using control points $\{g_j\}$ placed on a coarse grid in the normalized space. Each control point is assigned a random displacement vector
\begin{equation}
d_j = a_n \cdot \tilde{d}_j, \quad \tilde{d}_j \sim \mathcal{N}(0, I),
\label{eq:control_disp}
\end{equation}
where $a_n = a_{mm} / (\text{scale} + \epsilon)$ controls the deformation magnitude and $a_{mm} \sim \mathcal{U}(0, 15)$ mm.

For each surface point $s_i$, the displacement is computed via inverse-distance weighted interpolation from its $k$ nearest control points:
\begin{equation}
d(s_i) = \sum_{j \in \mathcal{N}_k(s_i)} w_{ij} d_j, 
\quad
w_{ij} = \frac{\|s_i - g_j\|^{-1}}{\sum_{l \in \mathcal{N}_k(s_i)} \|s_i - g_l\|^{-1}}.
\label{eq:interpolation}
\end{equation}

The final perturbed surface point is given by
\begin{equation}
\hat{s}_i = \tilde{s}_i + d(s_i).
\label{eq:final_surface}
\end{equation}
This formulation introduces realistic surface perturbations while preserving local geometric consistency. Noise is applied only to the observed partial surface $S$, while the pre-operative anatomy and ground-truth displacement field remain unchanged. Figure~\ref{fig:noise_levels} illustrates representative noisy surface samples under increasing noise levels and the corresponding distribution of per-point perturbation magnitudes.

\subsection{Evaluation Metrics}

Registration performance was evaluated using endpoint error (EPE),
root-mean-square error (RMSE). 

Let $\{\mathbf{x}_i\}_{i=1}^{N}$ denote the set of sampled points in the
preoperative space, $\mathbf{u}_i^{\mathrm{gt}}$ the corresponding
ground-truth displacement vectors, and $\mathbf{u}_i^{\mathrm{pred}}$
the predicted displacements.

\paragraph{Endpoint Error (EPE).}
The endpoint error measures the Euclidean distance between predicted and
ground-truth displacement vectors:

\begin{equation}
\mathrm{EPE} =
\frac{1}{N}\sum_{i=1}^{N}
\left\|
\mathbf{u}_i^{\mathrm{pred}} -
\mathbf{u}_i^{\mathrm{gt}}
\right\|_2 .
\end{equation}

\paragraph{Root-Mean-Square Error (RMSE).}
RMSE quantifies the global displacement error across all spatial
dimensions:

\begin{equation}
\mathrm{RMSE} =
\sqrt{
\frac{1}{3N}
\sum_{i=1}^{N}
\left\|
\mathbf{u}_i^{\mathrm{pred}} -
\mathbf{u}_i^{\mathrm{gt}}
\right\|_2^2
}.
\end{equation}

\subsection{Implementation Details and Training Settings}

The proposed network was implemented in PyTorch and trained on an NVIDIA
A100 GPU. 
The architecture employs a multi-scale decoder with $K=3$ hierarchical
levels, producing four displacement fields
$\{\mathbf{u}_k\}_{k=0}^{4}$.
These outputs are fused using fixed weights
$(\alpha_0,\alpha_1,\alpha_2,\alpha_3)=(1.0,0.3,0.3,0.05)$. For the number of points in the point cloud, an optimal threshold of 12000 points for the preoperative 6000 for the partial intraopertive surface was kept which allowed us train with different number of points without performing strict resampling. 

The model was optimized using the AdamW optimizer with an initial
learning rate of $1\times10^{-3}$ and a weight decay of $1\times10^{-4}$.
A cosine learning rate scheduler was employed to facilitate stable
convergence.
Training was conducted for 100 epochs with a batch size of 2.

Automatic mixed precision was used to accelerate training and reduce
memory consumption.
All network parameters were initialized using He initialization.
The total training time was approximately 37 hours per experiment.

The smoothness regularization weight was set to $\lambda_s=0.1$, which
was selected empirically based on validation performance.
\subsubsection{Baselines and Comparison Methods}
All baseline methods, including ICP, CPD, Livermatch, C2P-Net, and DefReg, were implemented using either publicly available official code or carefully reimplemented following the original publications when code was unavailable. For learning-based methods, training was performed on the same dataset using identical data splits and preprocessing as the proposed model to ensure a fair comparison. Hyperparameters were selected based on recommendations from the respective papers or tuned minimally for stable convergence. All methods were evaluated under the same experimental protocol.
\section{Results}

\subsection{Qualitative}
\begin{figure}[H]
  \centering
  \includegraphics[width=1.0\linewidth]{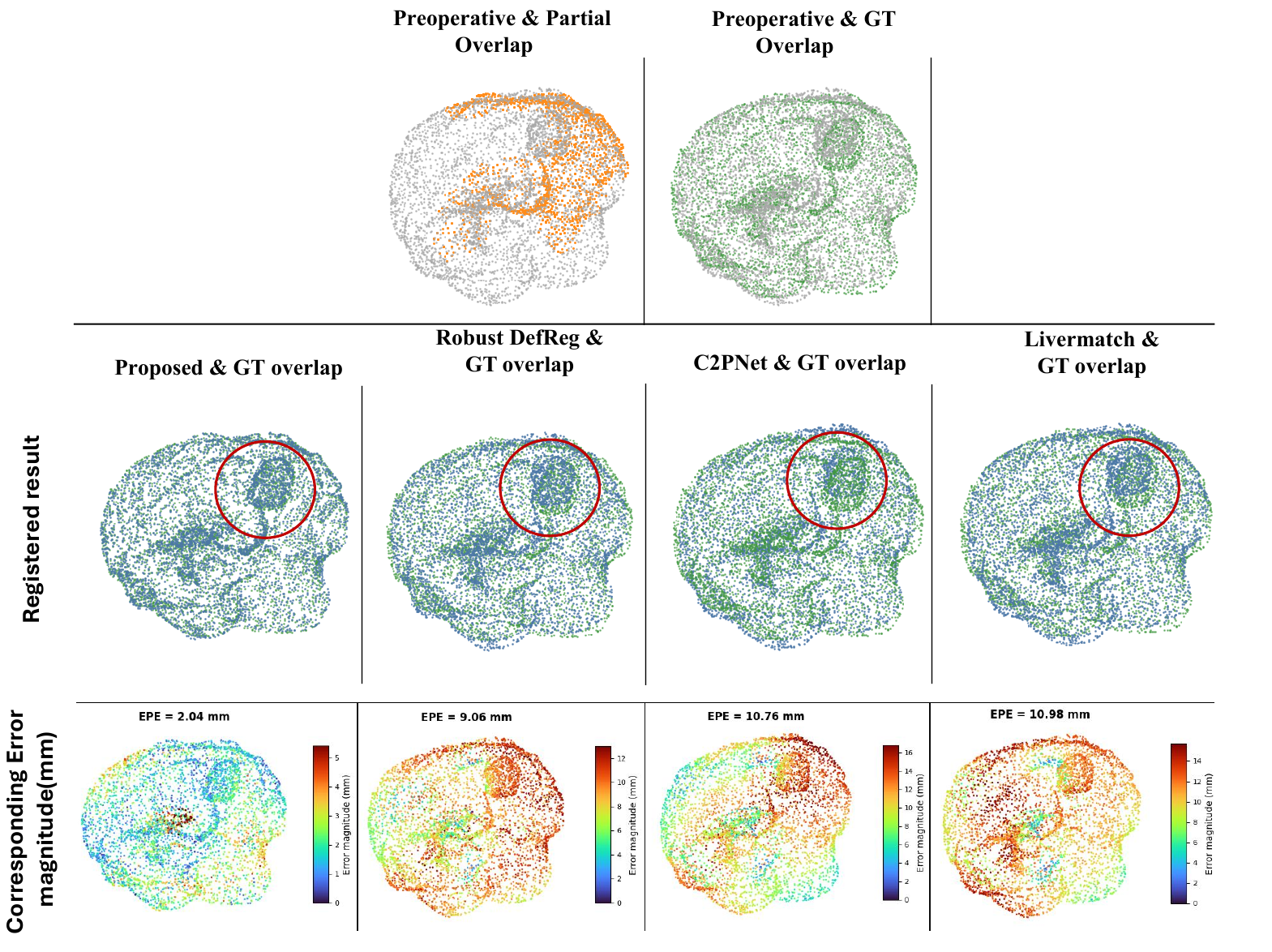}
  \caption{Qualitative comparison of registration performance across different learning based methods for Case 1.}
  \label{res5}
\end{figure}

\begin{figure}[H]
  \centering
  \includegraphics[width=1.0\linewidth]{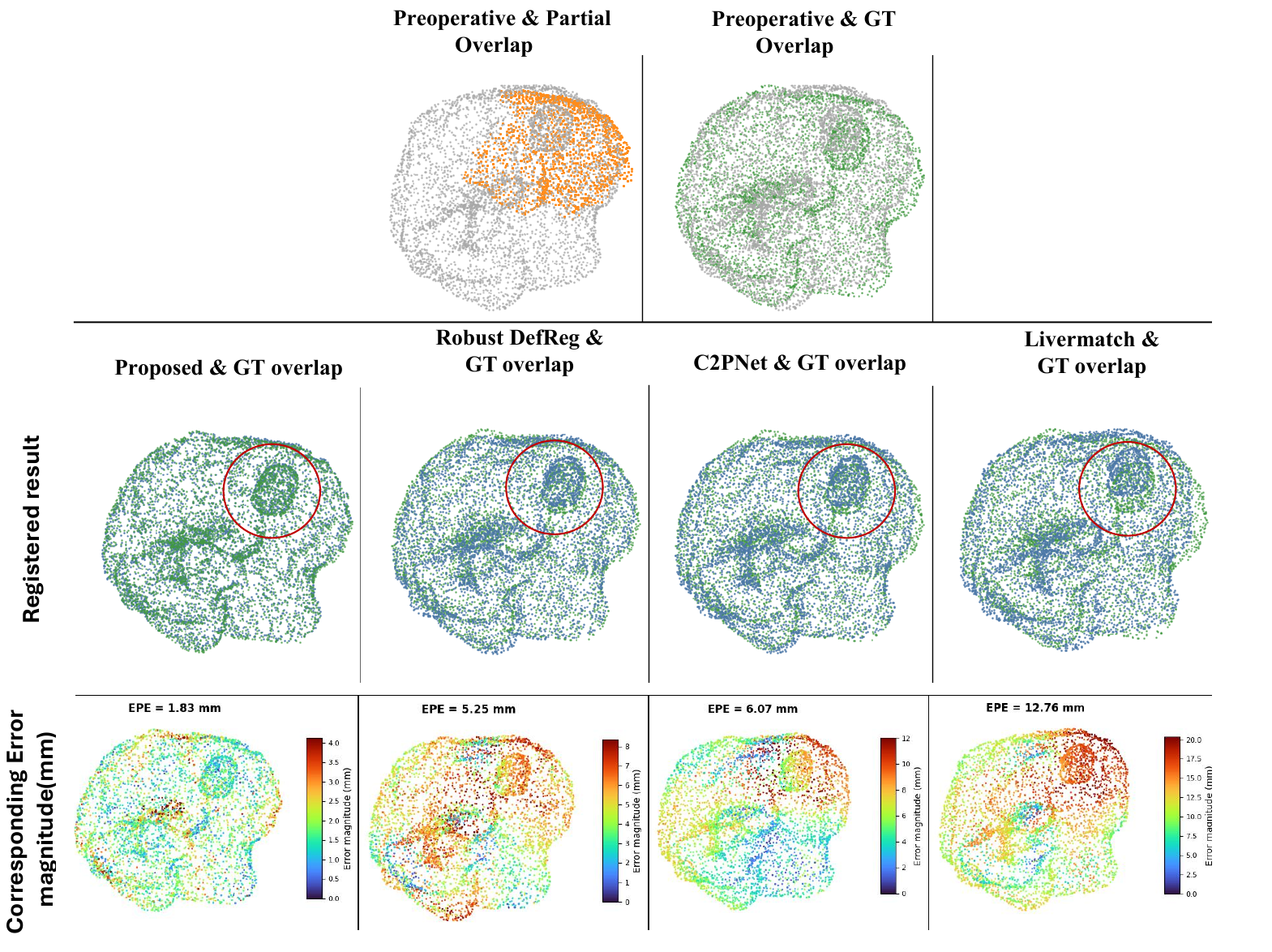}
  \caption{Qualitative comparison of registration performance across different learning based methods for case 2}
  \label{res6}
\end{figure}

Figs.~\ref{res5} and \ref{res6} present qualitative comparisons of different learning based registration methods under partial intraoperative surface observations for two representative cases. The first row shows the partial intraoperative surface (orange) overlaid on the preoperative (gray), and the Preoperative overlaid on the ground-truth intraoperative pointcloud. The subsequent rowa compare the outputs of the evaluated methods, with corresponding point-wise error maps (mm).

Across both cases, the proposed method consistently achieves more accurate alignment and lower error compared to the competing approaches. In Case 1 (Fig.~\ref{res5}), the proposed method attains an EPE of 2.04 mm, whereas alternative methods exhibit substantially higher errors, reaching up to 11.41 mm. A similar trend is observed in Case 2 (Fig.~\ref{res6}), where the proposed method maintains low error (1.83 mm), while other methods show increased misalignment, particularly in regions distant from the observed surface.

Qualitatively, the competing methods tend to produce noticeable deviations in anatomically complex regions (highlighted by the red circles), with errors propagating beyond the locally observed areas. In contrast, the proposed approach preserves the global structure more effectively and yields more coherent alignment across the entire surface. These results suggest that the proposed framework better captures global deformation patterns from partial observations, leading to improved robustness and generalization under limited intraoperative visibility.

\subsubsection{Impact of Intraoperative Surface Coverage and Spatial Distribution}
\begin{figure}[H]
  \centering
  \includegraphics[width=1.0\linewidth]{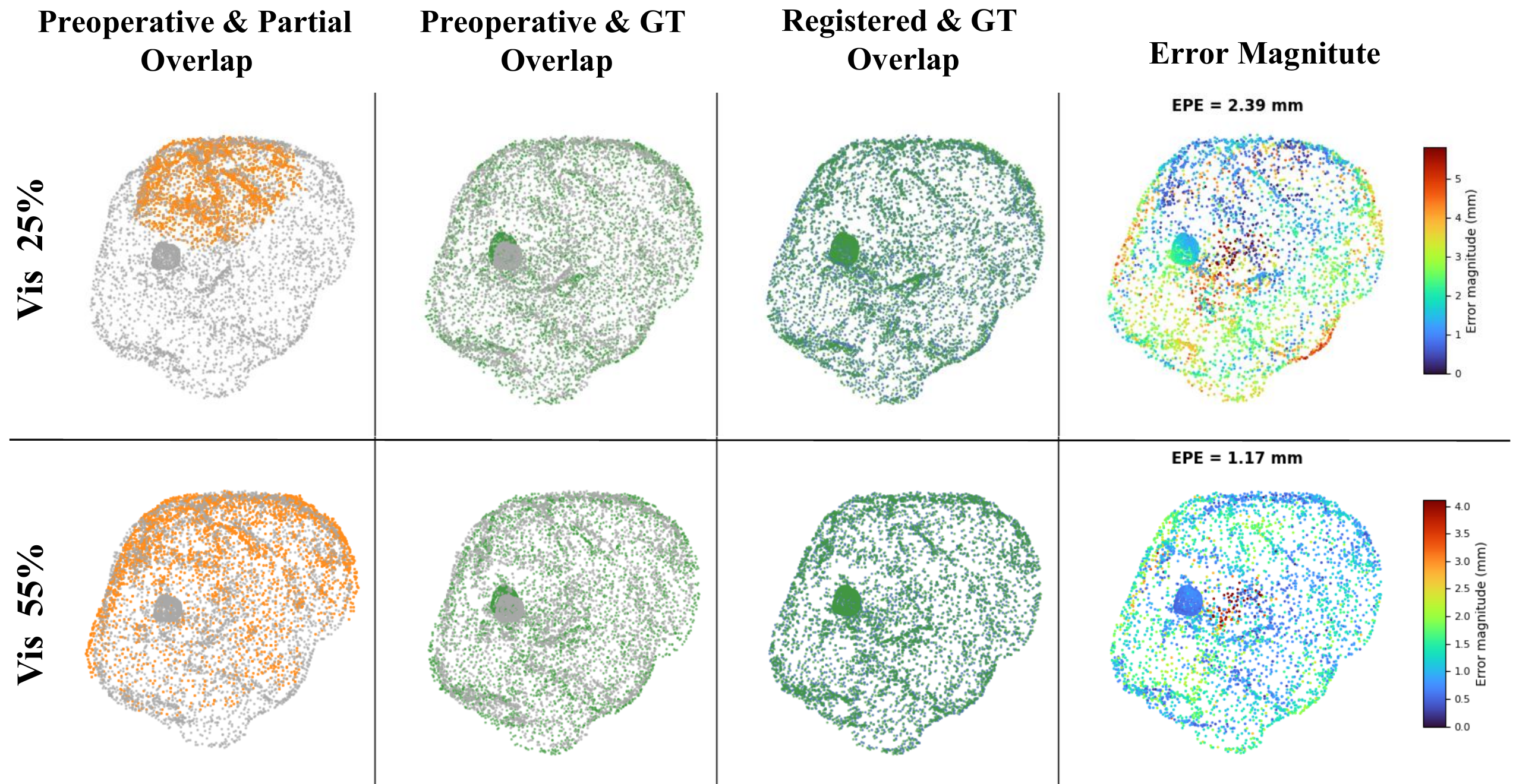}
  \caption{Registration result of Case 3: 25\% vs. 45\% visibility}
  \label{res1}
\end{figure}

\begin{figure}[H]
  \centering
  \includegraphics[width=1.0\linewidth]{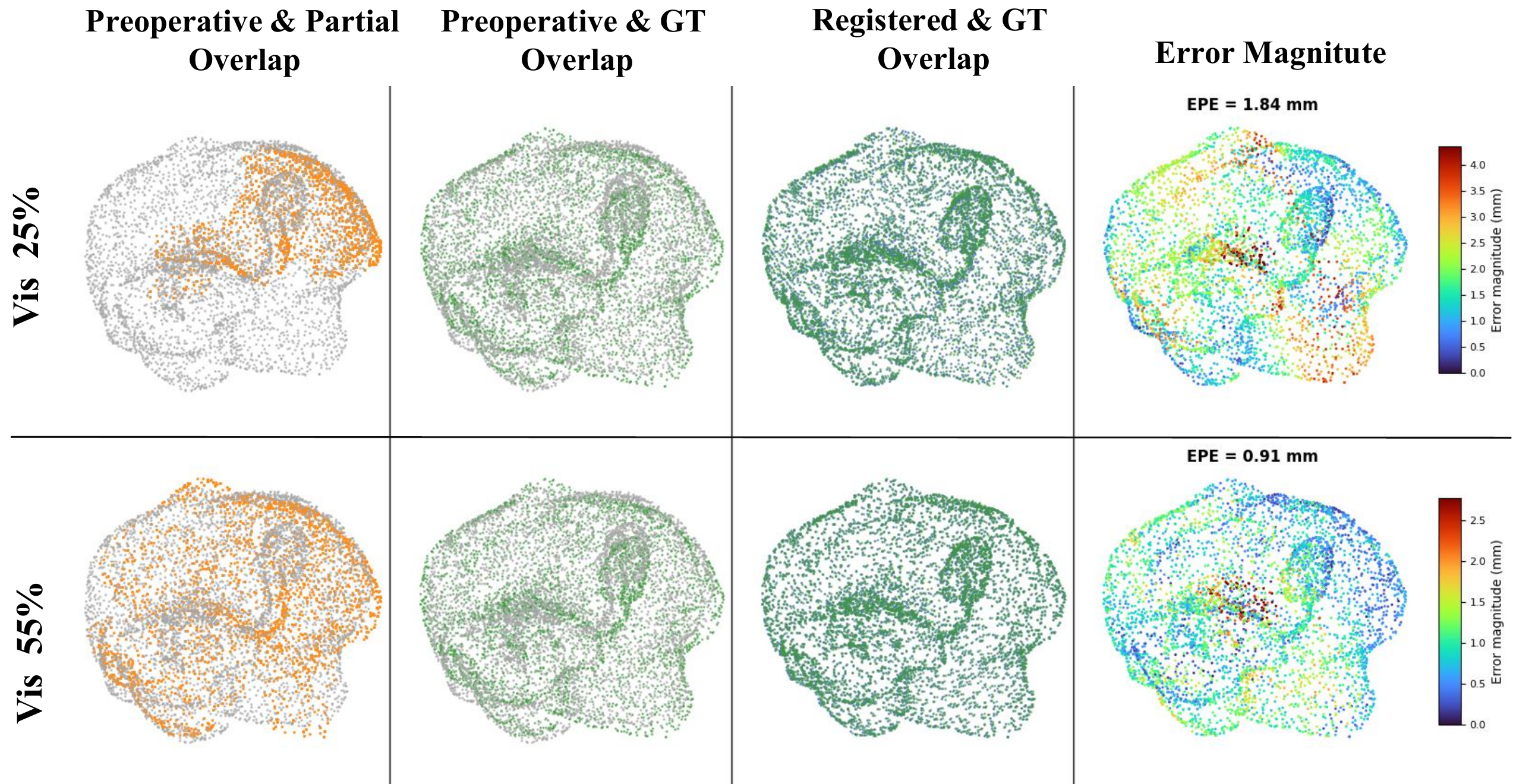}
  \caption{Registration result of Case 4: 25\% vs. 45\% visibility}
  \label{res2}
\end{figure}

\begin{figure}[H]
  \centering
  \includegraphics[width=1.0\linewidth]{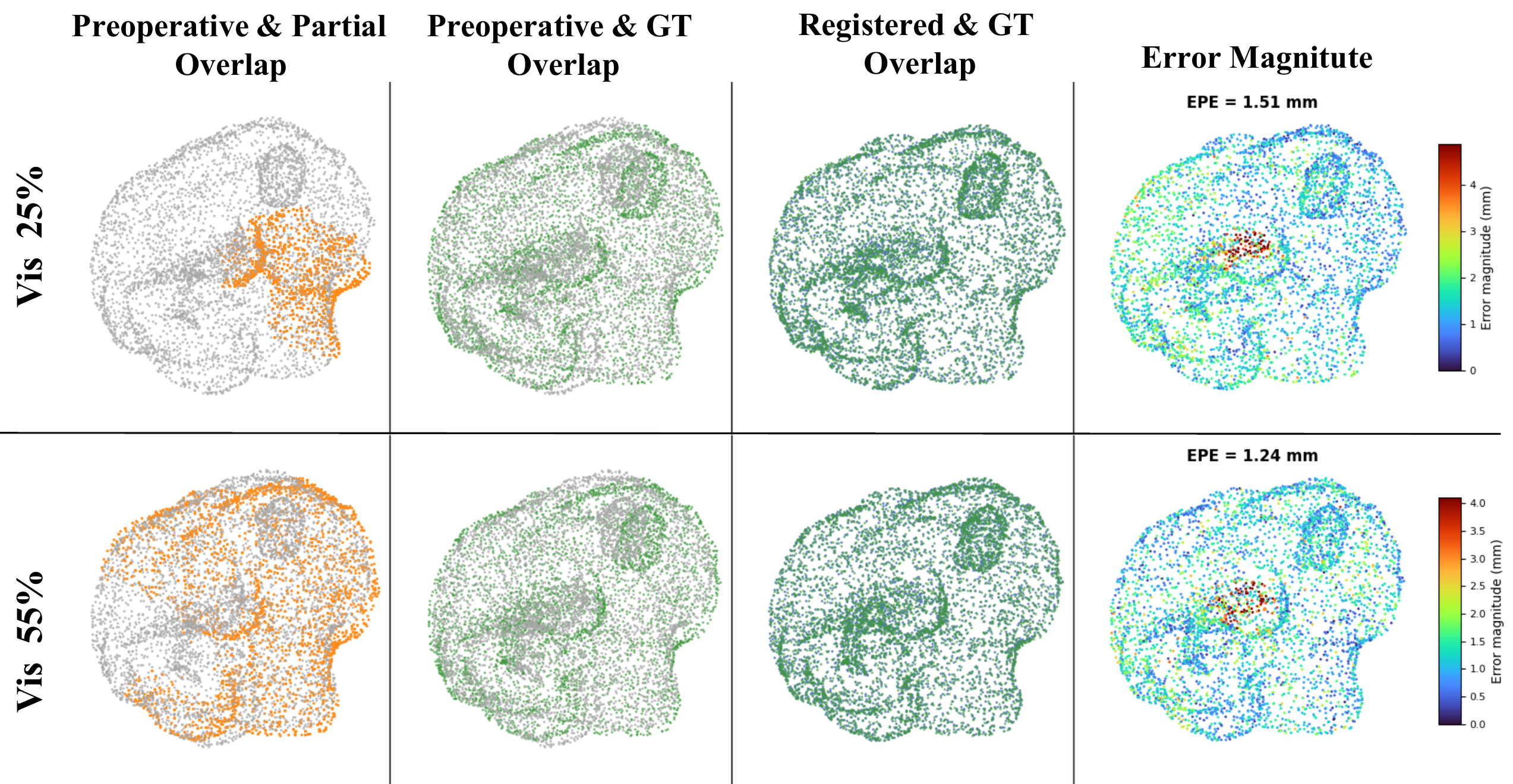}
  \caption{Registration result of Case 5: 25\% vs. 45\% visibility}
  \label{res3}
\end{figure}

The qualitative ablation in Fig.~\ref{res1}, \ref{res2}, \ref{res3} suggests that both visibility extent and spatial coverage influence registration performance, although not equally. In Cases 3, 4 and 4 increasing the visible surface from 25\% to 45\% results in improved alignment and lower error, indicating that additional intraoperative observations provide stronger geometric guidance for deformation estimation. However, Case 3 and 4 shows that accurate registration can still be achieved even when the visible surface does not predominantly cover the tumor-side or anticipated craniotomy region. This finding suggests that the proposed network is able to leverage global anatomical structure and learned deformation regularities, rather than relying solely on direct observations from the most clinically relevant local area. These results indicate that \emph{how much} of the intraoperative surface is seen generally matters, whereas \emph{where} it is seen may be less critical when sufficient global context is retained.

\begin{figure}[H]
  \centering
  \includegraphics[width=1.0\linewidth]{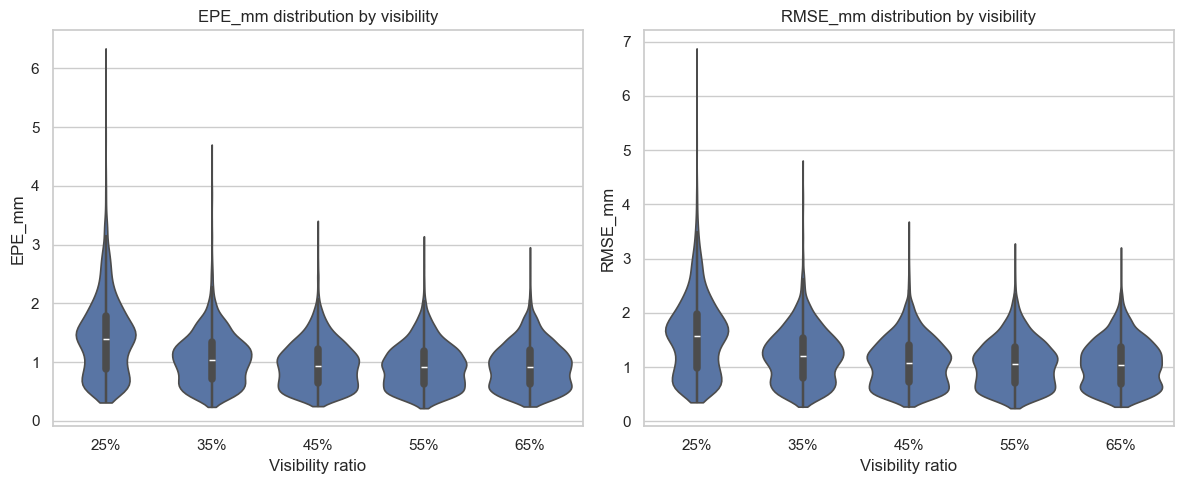}
  \caption{Distribution of registration errors under varying surface visibility. (a) Distribution of end-point error (EPE) across different visibility ratios. (b) Distribution of root mean square error (RMSE) between the registered preoperative and ground-truth intraoperative configurations. Each violin represents the error distribution over all test samples for a given visibility level (25–65\%). Increasing surface visibility leads to progressively reduced error and more stable registration performance.
  }
  \label{epe}
\end{figure}


\subsection{Quantitative Evaluation}
We quantitatively evaluated the registration performance of the proposed framework using the endpoint error (EPE) and root-mean-square error (RMSE), both reported in millimeters (mm). Lower values indicate better alignment accuracy. All results are reported as mean ± standard deviation over the test set.

Table \ref{tab:reg_error} presents a comparative evaluation of our method against several classical and learning-based baselines. Among all evaluated approaches, the proposed method achieves the lowest registration error in terms of both EPE and RMSE.


\begin{table}[H]
\centering
\caption{Quantitative comparison of registration accuracy in terms of
endpoint error (EPE) and root-mean-square error (RMSE).
Lower values indicate better performance.}
\label{tab:reg_error}
\renewcommand{\arraystretch}{1.5}
\begin{tabular}{lcc}
\toprule
\multicolumn{3}{c}{Initial misalignment} \\
\multicolumn{3}{c}{\emph{EPE: $8.13 \pm 4.47$ mm \quad RMSE: $7.42 \pm 3.64$ mm}} \\
\midrule
\midrule
Model & EPE (mm) & RMSE (mm) \\
\midrule
ICP \cite{besl1992method}       & $18.94 \pm 13.08$ & $21.20 \pm 10.09$ \\
CPD \cite{myronenko2010point}   & $16.76 \pm 8.34$  & $14.60 \pm 6.24$  \\
NICP \cite{amberg2007optimal}   & $15.57 \pm 3.51$  & $11.66\pm 1.89 $ \\
Livermatch \cite{yang2023learning} & $9.55  \pm 6.12$  & $10.69 \pm 7.43$  \\
C2P-Net \cite{liu2024non} & $6.87  \pm 3.14$  & $7.51  \pm 4.03$  \\
Robust DefReg \cite{monji2025robust}    & $4.65 \pm 1.98$  & $6.46 \pm 2.09$  \\
Proposed method & $\mathbf{1.13 \pm 0.75}$ & $\mathbf{1.33 \pm 0.81}$ \\
\bottomrule
\end{tabular}
\end{table}

Specifically, our approach obtains an EPE of 1.13 ± 0.75 mm and an RMSE of 1.33 ± 0.81 mm, outperforming all competing methods by a substantial margin. In contrast, traditional ICP exhibits limited performance due to its reliance on rigid or locally rigid correspondences, resulting in high residual errors. Similarly, learning-based baselines such as Livermatch, C2P-Net, DefReg, and demonstrate improved performance compared to ICP, but remain clearly inferior to the proposed framework.

The superior accuracy of our method can be attributed to its dedicated deformation modeling strategy and its ability to effectively learn the relationship between partial surface observations and volumetric displacement fields. Unlike existing approaches that primarily focus on surface-to-surface alignment, our framework directly optimizes volumetric consistency, leading to more accurate and stable registration results.



\subsubsection{Robustness to Surface Noise}

To assess the robustness of the proposed framework under noisy surface observations, we evaluated registration accuracy under different combinations of Gaussian noise and smoothing perturbations. The results are summarized in Table~\ref{tab:noise_perf}.

\begin{table}[H]
\centering
\caption{Registration accuracy under fixed surface noise levels (test set).}
\label{tab:noise_perf}
\setlength{\tabcolsep}{7pt}
\renewcommand{\arraystretch}{1.2}

\begin{tabular}{ccccc}
\toprule
$\sigma_{\mathrm{mm}}$ & $a_{\mathrm{mm}}$ & EPE (mm) $\downarrow$ & RMSE (mm) $\downarrow$  \\
\midrule
0   & 0   & $1.19 \pm ± 0.75$ & $1.33 \pm 0.81 $  \\
1   & 5   & $1.63 \pm 0.72 $ & $1.79 \pm 0.76 $ \\
2   & 10   & $2.23 \pm 0.85 $ & $2.39 \pm 0.91 $ \\
5   & 15   & $2.60 \pm 1.02 $ & $2.77 \pm 1.06 $ \\

\bottomrule
\end{tabular}
\end{table}

As shown in Table~\ref{tab:noise_perf}, the registration performance degrades gradually as the noise level increases. When no noise is applied ($\sigma_{\mathrm{mm}} = 0$, $\alpha_{\mathrm{mm}} = 0$), the model achieves an EPE of $1.19 \pm 0.75$~mm and an RMSE of $1.33 \pm 0.81$~mm. Even under the highest noise setting ($\sigma_{\mathrm{mm}} = 5$~mm, $\alpha_{\mathrm{mm}} = 15$~mm), the error remains limited, with an EPE of $2.60 \pm 1.02$~mm and an RMSE of $2.77 \pm 1.06$~mm.

This relatively small increase in error indicates that the proposed method is robust to surface perturbations. The gradual performance degradation suggests that the network learns stable geometric features and is not overly sensitive to local surface irregularities. This robustness is particularly important in neurosurgical scenarios, where intra-operative surface measurements are often affected by noise, occlusions, and reconstruction artifacts.


Table 3 compares the RMSE values of different methods under noisy surface conditions. While all baseline methods suffer noticeable performance degradation, the proposed framework consistently maintains the lowest error.

In particular, our method achieves an RMSE of $2.77 \pm 1.06$~mm, which is significantly lower than, Robust DefReg ($7.05 \pm 3.43$~mm). Traditional ICP and CPD also exhibits poor robustness, with an RMSE of $17.64 \pm 8.12$~mm and $23.16 \pm 13.56$~mm.

\begin{table}[H]
\centering
\caption{Comparison of RMSE (mm, Mean $\pm$ SD) of different algorithms under noisy surface observations.}
\small
\setlength{\tabcolsep}{4pt}
\renewcommand{\arraystretch}{1.15}
\begin{tabular}{lcc}
\toprule
\multicolumn{3}{c}{Initial misalignment\emph{\quad RMSE: $7.42 \pm 3.64$ mm}} \\
\midrule
Model & RMSE w/ noise & RMSE w/o noise \\
\midrule
ICP             & $23.16 \pm 13.56$ & $21.20 \pm 10.09$ \\
CPD         & $17.64 \pm 8.12$ & $14.60  \pm 6.24$  \\
Livermatch  & $10.23 \pm 5.31$ & $10.69  \pm 7.43$  \\
C2P-Net  & $8.38 \pm 6.29$ & $7.51  \pm 4.03$  \\
Robust DefReg          & $7.05 \pm 3.43$  & $6.46 \pm 2.09$  \\
Proposed method & $2.77 \pm 1.06$   & $1.33 \pm 0.81$   \\
\bottomrule
\end{tabular}

\vspace{2pt}
\end{table}
These results indicate that existing approaches struggle to handle incomplete and corrupted surface observations, often leading to unstable deformation estimates. In contrast, the proposed framework effectively leverages learned volumetric priors and multi-scale deformation modeling, allowing it to recover accurate displacement fields even under severe noise.

\subsubsection{Performance under different visibility Ratio}
Table~\ref{tab:rmse_compare} reports the registration performance of the proposed framework under varying levels of surface visibility. Different visibility ratios are considered to simulate realistic intra-operative conditions, where only partial and incomplete cortical surfaces are typically available due to limited surgical exposure, occlusions, and noise. As the visibility ratio increases, a consistent improvement in registration accuracy is observed, reflected by the monotonic decrease in EPE and RMSE values from 25\% to 65\% visibility. 

\begin{table}[H]
\centering
\caption{Comparison of registration performance the proposed framework under different surface visibility.}
\label{tab:rmse_compare}
\setlength{\tabcolsep}{8pt}
\renewcommand{\arraystretch}{1.2}

\begin{tabular}{lccc}
\toprule
Visibility ratio & EPE & RMSE \\
\midrule
 
 25\%     & $ 1.98 \pm 0.77$   & $2.01 \pm 0.83$ \\
 35\%     & $ 1.25 \pm 0.57$   & $1.42\pm 0.62 $  \\
 45\%     & $ 1.06 \pm 0.45$   & $1.21 \pm 0.51$ \\
 55\%     & $ 0.96 \pm 0.40$   & $1.09 \pm 0.45$ \\
 65\%     & $ 0.93 \pm 0.38$   & $1.06 \pm 0.43$  \\
\bottomrule 
\end{tabular}
\end{table}

At low visibility levels (25\%--35\%), the proposed method maintains stable performance despite severely limited surface observations, indicating its ability to infer volumetric deformations from sparse geometric cues. Further improvements are achieved at intermediate visibility ratios (45\%--55\%), where additional surface information provides stronger constraints for deformation estimation. At higher visibility levels ($\geq$55\%), the errors approach < 1 mm , demonstrating near-perfect alignment when sufficient cortical coverage is available. 
\subsection{Real-Time Capability (Frame Rate)}

The computational efficiency of the proposed framework was evaluated in terms of the average processing time per sample and the corresponding frames per second (FPS). For each input pair $(V, S)$, the total runtime consists of two components: (i) the forward pass of the network to predict the dense deformation vector field $\phi$, and (ii) the subsequent deformation update applied to the preoperative point cloud. The total runtime is defined as:
\begin{equation}
T_{\text{total}} = T_{\text{inference}} + T_{\text{update}}
\end{equation}
where $T_{\text{inference}}$ denotes the time required for a single forward pass of the network, and $T_{\text{update}}$ corresponds to the point-wise deformation step:
\begin{equation}
V_{\text{reg}} = V + \phi
\end{equation}

To facilitate comparison with existing methods, the runtime is additionally reported in terms of frames per second (FPS), computed as:
\begin{equation}
\text{FPS} = \frac{1000}{T_{\text{total}}}
\end{equation}
where $T_{\text{total}}$ is expressed in milliseconds.
\begin{table}[H]
\centering
\caption{Runtime analysis of the proposed framework. The end-to-end time includes both inference time corresponding one forward pass to predict the deformation vector field followed by registration $(V,S) \rightarrow {\phi} \rightarrow V_{\text{reg}} = V +{\phi}$.} 
\label{tab:runtime_breakdown}
\setlength{\tabcolsep}{8pt}
\renewcommand{\arraystretch}{1.2}

\begin{tabular}{lcc}
\toprule
Method &  Time Per sample(ms) $\downarrow$ & FPS $\uparrow$ \\
\midrule
CPD & 75288.15 & 0.1 \\
ICP & 216.99 & 4.61 \\
Non-rigid ICP & 414.40 &  2.41 \\ 
Robust DefReg & 861.73 & 1.16 \\
\textbf{Proposed end-to-end update} & \textbf{175.28} & \textbf {5.71}\\
\bottomrule
\end{tabular}
\end{table}
All measurements were conducted on an NVIDIA A100 GPU. Classical baselines (ICP, CPD, and their non-rigid variants) were evaluated using their standard implementations under comparable settings.


\section{Discussion \& Conclusion}
This study addresses the problem of estimating dense volumetric deformation from sparse and partial intra-operative surface observations. In contrast to conventional image-to-image registration methods that rely on dense intra-operative imaging, the proposed framework learns to infer displacement vector fields directly from limited cortical surface measurements. This formulation reflects realistic neurosurgical constraints, where only partial surface visibility is available and repeated volumetric imaging may disrupt workflow.

The quantitative results demonstrate consistent improvements over both classical and learning-based baselines. The proposed method achieves an EPE of $1.13 \pm 0.75$ mm and an RMSE of $1.33 \pm 0.81$ mm, outperforming recent point-based learning approaches by a substantial margin. Traditional ICP exhibits high residual errors due to its sensitivity to initialization and its inability to model complex non-rigid deformation. Although learning-based baselines improve performance compared to ICP and CPD, they primarily focus on surface alignment and do not explicitly model volumetric deformation consistency. In contrast, the proposed framework directly supervises dense displacement fields, enabling more stable and anatomically coherent predictions.

The proposed method also demonstrates favorable computational efficiency. The end-to-end runtime of 175.28 ms per sample corresponds to approximately 5.71 FPS. In contrast, classical methods such as CPD and non-rigid ICP require iterative optimization, resulting in significantly higher runtimes. The single forward-pass inference combined with a direct deformation update allows the proposed framework to maintain both accuracy and computational efficiency, which is critical for intra-operative deployment.

The robustness analysis under different surface noise conditions further supports the stability of the proposed approach. As noise levels increase, registration accuracy degrades gradually rather than abruptly. Even under the highest noise setting, the error increase remains moderate. This behavior suggests that the network does not overfit to precise surface correspondences but instead learns deformation priors that are resilient to local surface irregularities and spatially coherent perturbations. Such robustness is essential in neurosurgical settings, where intra-operative surface acquisition is inherently affected by measurement uncertainty, occlusion, and reconstruction artifacts.

The visibility experiments reveal a clear relationship between surface coverage and deformation accuracy. As the visibility ratio increases from 25\% to 65\%, both EPE and RMSE decrease monotonically. However, the model maintains stable performance even at low visibility levels, indicating that it can infer volumetric deformation from sparse geometric cues. This ability is particularly important for brain shift compensation, where surgical exposure is limited and asymmetric. The hierarchical decoder and multi-scale feature aggregation likely contribute to this behavior by combining global structural context with fine-scale geometric reasoning.

Compared to previously reported volume-to-surface learning approaches in other anatomical regions, brain shift presents additional challenges due to heterogeneous deformation patterns and severely restricted surface visibility. The proposed framework explicitly addresses this asymmetry between a complete pre-operative model and a partial intra-operative observation. By integrating multi-scale feature extraction with progressive deformation refinement, the method demonstrates that volumetric deformation recovery from sparse cortical data is feasible without reliance on intra-operative volumetric imaging.

Despite these promising results, several limitations should be acknowledged. The deformation fields used for training are generated synthetically using analytic operators. While designed to approximate plausible intra-operative patterns, they may not capture the full biomechanical complexity of real brain shift. Furthermore, evaluation is conducted on simulated partial surfaces derived from MRI-based models rather than real intra-operative stereo or LRS acquisitions. The current framework also does not explicitly enforce biomechanical constraints such as incompressibility or diffeomorphic regularity. Future work should explore integration of physics-informed priors, validation on real intra-operative surface datasets, and clinical studies assessing navigation improvement.

Overall, the findings indicate that learning volumetric deformation directly from sparse surface observations is a viable and effective strategy for brain shift compensation. The proposed framework offers a workflow-compatible alternative to intra-operative volumetric imaging and provides a scalable foundation for near real-time soft-tissue deformation tracking in image-guided neurosurgery.
\vspace{1.9em}

\textbf{Acknowledgement:}
This study has been supported by the ACMIT COMET Competence Center (FFG project number: 879733, application number: 28997164) and the COMET Module SD-OpT (FFG project number: FO999888360, application number: 39955962), both funded by the federal government of Austria and the governments of the provinces Lower Austria and Tyrol. In addition, Ander Biguri acknowledges the Accelerate Programme for Scientific Discovery.







\bibliographystyle{elsarticle-num-names} 
\bibliography{references}

@article{kelly1986computer,
  title={Computer-assisted stereotaxic laser resection of intra-axial brain neoplasms},
  author={Kelly, Patrick J and Kall, Bruce A and Goerss, Stephan and Earnest, Franklin},
  journal={Journal of neurosurgery},
  volume={64},
  number={3},
  pages={427--439},
  year={1986},
  publisher={Journal of Neurosurgery Publishing Group}
}

@article{hastreiter2004strategies,
  title={Strategies for brain shift evaluation},
  author={Hastreiter, Peter and Rezk-Salama, Christof and Soza, Grzegorz and Bauer, Michael and Greiner, G{\"u}nther and Fahlbusch, Rudolf and Ganslandt, Oliver and Nimsky, Christopher},
  journal={Medical image analysis},
  volume={8},
  number={4},
  pages={447--464},
  year={2004},
  publisher={Elsevier}
}

@article{xiao2019evaluation,
  title={Evaluation of MRI to ultrasound registration methods for brain shift correction: the CuRIOUS2018 challenge},
  author={Xiao, Yiming and Rivaz, Hassan and Chabanas, Matthieu and Fortin, Maryse and Machado, Ines and Ou, Yangming and Heinrich, Mattias P and Schnabel, Julia A and Zhong, Xia and Maier, Andreas and others},
  journal={IEEE transactions on medical imaging},
  volume={39},
  number={3},
  pages={777--786},
  year={2019},
  publisher={IEEE}
}

@article{zhang2019wireless,
  title={Wireless transmission-based brain shift compensation system},
  author={Zhang, Chenxi and Dong, Yuan},
  journal={The Journal of Engineering},
  volume={2019},
  number={14},
  pages={506--511},
  year={2019},
  publisher={Wiley Online Library}
}

@inproceedings{sinha2003cortical,
  title={Cortical shift tracking using a laser range scanner and deformable registration methods},
  author={Sinha, Tuhin K and Duay, Valerie and Dawant, Benoit M and Miga, Michael I},
  booktitle={International Conference on Medical Image Computing and Computer-Assisted Intervention},
  pages={166--174},
  year={2003},
  organization={Springer}
}

@inproceedings{zhang2024comprehensive,
  title={A Comprehensive Survey and Taxonomy on Point Cloud Registration Based on Deep Learning.},
  author={Zhang, Yu-Xin and Gui, Jie and Cong, Xiaofeng and Gong, Xin and Tao, Wenbing},
  booktitle={IJCAI},
  pages={8344--8353},
  year={2024}
}

@article{yang2017stereovision,
  title={Stereovision-based integrated system for point cloud reconstruction and simulated brain shift validation},
  author={Yang, Xiaochen and Clements, Logan W and Luo, Ma and Narasimhan, Saramati and Thompson, Reid C and Dawant, Benoit M and Miga, Michael I},
  journal={Journal of Medical Imaging},
  volume={4},
  number={3},
  pages={035002--035002},
  year={2017},
  publisher={Society of Photo-Optical Instrumentation Engineers}
}

@article{correa2017neurosurgery,
  title={Neurosurgery and brain shift: review of the state of the art and main contributions of robotics},
  author={Correa-Arana, Karin and Vivas-Alb{\'a}n, Oscar A and Sabater-Navarro, Jos{\'e} M},
  journal={TecnoL{\'o}gicas},
  volume={20},
  number={40},
  pages={125--138},
  year={2017},
  publisher={Instituto Tecnol{\'o}gico Metropolitano-ITM}
}

@article{leroy2020intraoperative,
  title={Intraoperative MRI guidance for right deep fronto-temporal glioma resection: how I do it},
  author={Leroy, Henri-Arthur and Tuleasca, Constantin and Vannod-Michel, Quentin and Reyns, Nicolas},
  journal={Acta Neurochirurgica},
  volume={162},
  number={12},
  pages={3037--3041},
  year={2020},
  publisher={Springer}
}

@article{clatz2005robust,
  title={Robust nonrigid registration to capture brain shift from intraoperative MRI},
  author={Clatz, Olivier and Delingette, Herv{\'e} and Talos, I-F and Golby, Alexandra J and Kikinis, Ron and Jolesz, Ferenc A and Ayache, Nicholas and Warfield, Simon K},
  journal={IEEE transactions on medical imaging},
  volume={24},
  number={11},
  pages={1417--1427},
  year={2005},
  publisher={IEEE}
}

@article{schulz2012intraoperative,
  title={Intraoperative image guidance in neurosurgery: development, current indications, and future trends},
  author={Schulz, Chris and Waldeck, Stephan and Mauer, Uwe Max},
  journal={Radiology research and practice},
  volume={2012},
  number={1},
  pages={197364},
  year={2012},
  publisher={Wiley Online Library}
}

@article{fu2020deep,
  title={Deep learning in medical image registration: a review},
  author={Fu, Yabo and Lei, Yang and Wang, Tonghe and Curran, Walter J and Liu, Tian and Yang, Xiaofeng},
  journal={Physics in Medicine \& Biology},
  volume={65},
  number={20},
  pages={20TR01},
  year={2020},
  publisher={IOP Publishing}
}

@article{de2019deep,
  title={A deep learning framework for unsupervised affine and deformable image registration},
  author={De Vos, Bob D and Berendsen, Floris F and Viergever, Max A and Sokooti, Hessam and Staring, Marius and I{\v{s}}gum, Ivana},
  journal={Medical image analysis},
  volume={52},
  pages={128--143},
  year={2019},
  publisher={Elsevier}
}

@article{yang2023learning,
  title={Learning feature descriptors for pre-and intra-operative point cloud matching for laparoscopic liver registration},
  author={Yang, Zixin and Simon, Richard and Linte, Cristian A},
  journal={International journal of computer assisted radiology and surgery},
  volume={18},
  number={6},
  pages={1025--1032},
  year={2023},
  publisher={Springer}
}

@misc{dalca2018unsupervised,
  title={Unsupervised Learning for Fast Probabilistic Diffeomorphic Registration, 729--738},
  author={Dalca Adrian, V and Guha, Balakrishnan and John, Guttag and Sabuncu Mert, R},
  year={2018},
  publisher={Springer International Publishing}
}

@article{balakrishnan2019voxelmorph,
  title={Voxelmorph: a learning framework for deformable medical image registration},
  author={Balakrishnan, Guha and Zhao, Amy and Sabuncu, Mert R and Guttag, John and Dalca, Adrian V},
  journal={IEEE transactions on medical imaging},
  volume={38},
  number={8},
  pages={1788--1800},
  year={2019},
  publisher={IEEE}
}

@misc{meshmixer,
  title        = {Autodesk Meshmixer},
  author       = {{Autodesk Inc.}},
  year         = {2024},
  howpublished = {\url{https://www.meshmixer.com}},
  note         = {Accessed: 2026-02-18}
}

@article{3d,
  title     = {3D Slicer as an image computing platform for the Quantitative Imaging Network},
  author    = {Fedorov, Andriy and Beichel, Reinhard and Kalpathy-Cramer, Jayashree and Finet, Julien and Fillion-Robin, Jean-Christophe and Pujol, Sonia and Bauer, Christian and Jennings, Dominique and Fennessy, Fiona and Sonka, Milan and others},
  journal   = {Magnetic Resonance Imaging},
  volume    = {30},
  number    = {9},
  pages     = {1323--1341},
  year      = {2012},
  publisher = {Elsevier}
}

@article{qi2017pointnetplusplus,
  title={PointNet++: Deep Hierarchical Feature Learning on Point Sets in a Metric Space},
  author={Qi, Charles R and Yi, Li and Su, Hao and Guibas, Leonidas J},
  journal={Advances in Neural Information Processing Systems},
  volume={30},
  year={2017}
}

@article{pereira2016volumetric,
  title={Volumetric measurements of brain shift using intraoperative cone-beam computed tomography: preliminary study},
  author={Pereira, Vitor Mendes and Smit-Ockeloen, Iris and Brina, Olivier and Babic, Drazenko and Breeuwer, Marcel and Schaller, Karl and Lovblad, Karl-Olof and Ruijters, Daniel},
  journal={Operative Neurosurgery},
  volume={12},
  number={1},
  pages={4--13},
  year={2016},
  publisher={LWW}
}

@article{liao2010automatic,
  title={Automatic recognition of midline shift on brain CT images},
  author={Liao, Chun-Chih and Xiao, Furen and Wong, Jau-Min and Chiang, I-Jen},
  journal={Computers in biology and medicine},
  volume={40},
  number={3},
  pages={331--339},
  year={2010},
  publisher={Elsevier}
}

@ARTICLE{6193441,
  author={DeLorenzo, Christine and Papademetris, Xenophon and Staib, Lawrence H. and Vives, Kenneth P. and Spencer, Dennis D. and Duncan, James S.},
  journal={IEEE Transactions on Medical Imaging}, 
  title={Volumetric Intraoperative Brain Deformation Compensation: Model Development and Phantom Validation}, 
  year={2012},
  volume={31},
  number={8},
  pages={1607-1619},
  doi={10.1109/TMI.2012.2197407}}

@inproceedings{haker2004landmark,
  title={Landmark-guided surface matching and volumetric warping for improved prostate biopsy targeting and guidance},
  author={Haker, Steven and Warfield, Simon K and Tempany, Clare MC},
  booktitle={International Conference on Medical Image Computing and Computer-Assisted Intervention},
  pages={853--861},
  year={2004},
  organization={Springer}
}

@inproceedings{taubin1995curve,
  title={Curve and surface smoothing without shrinkage},
  author={Taubin, Gabriel},
  booktitle={Proceedings of IEEE international conference on computer vision},
  pages={852--857},
  year={1995},
  organization={IEEE}
}

@phdthesis{prananta2010robotic,
  title={A robotic needle guide for prostate brachytherapy with pre-operative to intra-operative prostate volumes registration},
  author={Prananta, Thomas Diego},
  year={2010},
  school={University of British Columbia}
}

@article{henrich2025ludo,
  title={Ludo: Low-latency understanding of deformable objects using point cloud occupancy functions},
  author={Henrich, Pit and Mathis-Ullrich, Franziska and Scheikl, Paul Maria},
  journal={IEEE Transactions on Robotics},
  year={2025},
  publisher={IEEE}
}

@ARTICLE{10304319,

  author={Lin, Qinyong and Guo, Xiongbo and Xie, Yangjie and Peng, Kehai and Yang, Rongqian and Cai, Ken},

  journal={IEEE Transactions on Consumer Electronics}, 

  title={Surface Matching-Based Markerless Global Optimization Registration for Improved Optical Surgical Systems in Internet of Things-Enabled Operating Rooms}, 

  year={2024},

  volume={70},

  number={1},

  pages={939-946},

  doi={10.1109/TCE.2023.3329032}}

@article{wang2024brainmorph,
  title={BrainMorph: A foundational keypoint model for robust and flexible brain MRI registration},
  author={Wang Alan, Q and Rachit, Saluja and Heejong, Kim and He, Dalca Adrian and Sabuncu Mert, R},
  journal={arXiv preprint arXiv:2405.14019},
  year={2024}
}

@article{fan2020robust,
  title={A robust automated surface-matching registration method for neuronavigation},
  author={Fan, Yifeng and Yao, Xufeng and Xu, Xiufang},
  journal={Medical physics},
  volume={47},
  number={7},
  pages={2755--2767},
  year={2020},
  publisher={Wiley Online Library}
}

@article{liu2024non,
  title={Non-rigid point cloud registration for middle ear diagnostics with endoscopic optical coherence tomography},
  author={Liu, Peng and Golde, Jonas and Morgenstern, Joseph and Bodenstedt, Sebastian and Li, Chenpan and Hu, Yujia and Chen, Zhaoyu and Koch, Edmund and Neudert, Marcus and Speidel, Stefanie},
  journal={International Journal of Computer Assisted Radiology and Surgery},
  volume={19},
  number={1},
  pages={139--145},
  year={2024},
  publisher={Springer}
}

@article{geuzaine2009gmsh,
  title   = {Gmsh: A 3-D finite element mesh generator with built-in pre- and post-processing facilities},
  author  = {Geuzaine, Christophe and Remacle, Jean-François},
  journal = {International Journal for Numerical Methods in Engineering},
  volume  = {79},
  number  = {11},
  pages   = {1309--1331},
  year    = {2009},
  doi     = {10.1002/nme.2579}
}

@inproceedings{besl1992method,
  title={Method for registration of 3-D shapes},
  author={Besl, Paul J and McKay, Neil D},
  booktitle={Sensor fusion IV: control paradigms and data structures},
  volume={1611},
  pages={586--606},
  year={1992},
  organization={Spie}
}

@article{juvekar2024remind,
  title={Remind: The brain resection multimodal imaging database},
  author={Juvekar, Parikshit and Dorent, Reuben and K{\"o}gl, Fryderyk and Torio, Erickson and Barr, Colton and Rigolo, Laura and Galvin, Colin and Jowkar, Nick and Kazi, Anees and Haouchine, Nazim and others},
  journal={Scientific Data},
  volume={11},
  number={1},
  pages={494},
  year={2024},
  publisher={Nature Publishing Group UK London}
}

@inproceedings{li2019deepgcns,
  title={DeepGCNs: Can GCNs go as deep as CNNs?},
  author={Li, Guohao and Muller, Matthias and Thabet, Ali and Ghanem, Bernard},
  booktitle={Proceedings of the IEEE/CVF international conference on computer vision},
  pages={9267--9276},
  year={2019}
}

@article{falta2023lung250m,
  title={Lung250M-4B: a combined 3D dataset for CT-and point cloud-based intra-patient lung registration},
  author={Falta, Fenja and Gro{\ss}br{\"o}hmer, Christoph and Hering, Alessa and Bigalke, Alexander and Heinrich, Mattias},
  journal={Advances in Neural Information Processing Systems},
  volume={36},
  pages={54819--54832},
  year={2023}
}

@inproceedings{huang2021predator,
  title={Predator: Registration of 3d point clouds with low overlap},
  author={Huang, Shengyu and Gojcic, Zan and Usvyatsov, Mikhail and Wieser, Andreas and Schindler, Konrad},
  booktitle={Proceedings of the IEEE/CVF Conference on computer vision and pattern recognition},
  pages={4267--4276},
  year={2021}
}

@inproceedings{miga2000model,
  title={Model-updated image-guided neurosurgery: Preliminary analysis using intraoperative MR},
  author={Miga, Michael I and Staubert, Andreas and Paulsen, Keith D and Kennedy, Francis E and Tronnier, Volker M and Roberts, David W and Hartov, Alex and Platenik, Leah A and Lunn, Karen E},
  booktitle={International Conference on Medical Image Computing and Computer-Assisted Intervention},
  pages={115--124},
  year={2000},
  organization={Springer}
}

@inproceedings{aoki2019pointnetlk,
  title={Pointnetlk: Robust \& efficient point cloud registration using pointnet},
  author={Aoki, Yasuhiro and Goforth, Hunter and Srivatsan, Rangaprasad Arun and Lucey, Simon},
  booktitle={Proceedings of the IEEE/CVF conference on computer vision and pattern recognition},
  pages={7163--7172},
  year={2019}
}

@article{phan2018dgcnn,
  title={Dgcnn: A convolutional neural network over large-scale labeled graphs},
  author={Phan, Anh Viet and Le Nguyen, Minh and Nguyen, Yen Lam Hoang and Bui, Lam Thu},
  journal={Neural Networks},
  volume={108},
  pages={533--543},
  year={2018},
  publisher={Elsevier}
}

@inproceedings{qi2017pointnet,
  title={Pointnet: Deep learning on point sets for 3d classification and segmentation},
  author={Qi, Charles R and Su, Hao and Mo, Kaichun and Guibas, Leonidas J},
  booktitle={Proceedings of the IEEE conference on computer vision and pattern recognition},
  pages={652--660},
  year={2017}
}

@article{pfeiffer2019learning,
  title={Learning soft tissue behavior of organs for surgical navigation with convolutional neural networks},
  author={Pfeiffer, Micha and Riediger, Carina and Weitz, J{\"u}rgen and Speidel, Stefanie},
  journal={International journal of computer assisted radiology and surgery},
  volume={14},
  number={7},
  pages={1147--1155},
  year={2019},
  publisher={Springer}
}

@inproceedings{amberg2007optimal,
  title={Optimal step nonrigid ICP algorithms for surface registration},
  author={Amberg, Brian and Romdhani, Sami and Vetter, Thomas},
  booktitle={2007 IEEE conference on computer vision and pattern recognition},
  pages={1--8},
  year={2007},
  organization={IEEE}
}

@article{yang2025resolving,
  title={Resolving the ambiguity of complete-to-partial point cloud registration for image-guided liver surgery with patches-to-partial matching},
  author={Yang, Zixin and Heiselman, Jon S and Han, Cheng and Merrell, Kelly and Simon, Richard and Linte, Cristian A},
  journal={IEEE Journal of Biomedical and Health Informatics},
  year={2025},
  publisher={IEEE}
}

@article{gerard2017brain,
  title={Brain shift in neuronavigation of brain tumors: A review},
  author={Gerard, Ian J and Kersten-Oertel, Marta and Petrecca, Kevin and Sirhan, Denis and Hall, Jeffery A and Collins, D Louis},
  journal={Medical image analysis},
  volume={35},
  pages={403--420},
  year={2017},
  publisher={Elsevier}
}

@article{mitsui2011skin,
  title={Skin shift and its effect on navigation accuracy in image-guided neurosurgery},
  author={Mitsui, Takashi and Fujii, Masazumi and Tsuzaka, Masatoshi and Hayashi, Yuichiro and Asahina, Yoshinori and Wakabayashi, Toshihiko},
  journal={Radiological physics and technology},
  volume={4},
  number={1},
  pages={37--42},
  year={2011},
  publisher={Springer}
}

@article{hill1998measurement,
  title={Measurement of intraoperative brain surface deformation under a craniotomy},
  author={Hill, Derek LG and Maurer Jr, Calvin R and Maciunas, Robert J and Barwise, John A and Fitzpatrick, Michael J and Wang, Matthew Y},
  journal={Neurosurgery},
  volume={43},
  number={3},
  pages={514--526},
  year={1998},
  publisher={LWW}
}

@article{hammoud1996use,
  title={Use of intraoperative ultrasound for localizing tumors and determining the extent of resection: a comparative study with magnetic resonance imaging},
  author={Hammoud, Maarouf A and Ligon, B Lee and Elsouki, Rabih and Shi, Wei Ming and Schomer, Donald F and Sawaya, Raymond},
  journal={Journal of neurosurgery},
  volume={84},
  number={5},
  pages={737--741},
  year={1996},
  publisher={Journal of Neurosurgery Publishing Group}
}

@article{bayer2017intraoperative,
  title={Intraoperative imaging modalities and compensation for brain shift in tumor resection surgery},
  author={Bayer, Siming and Maier, Andreas and Ostermeier, Martin and Fahrig, Rebecca},
  journal={International journal of biomedical imaging},
  volume={2017},
  number={1},
  pages={6028645},
  year={2017},
  publisher={Wiley Online Library}
}

@article{myronenko2010point,
  title={Point set registration: Coherent point drift},
  author={Myronenko, Andriy and Song, Xubo},
  journal={IEEE transactions on pattern analysis and machine intelligence},
  volume={32},
  number={12},
  pages={2262--2275},
  year={2010},
  publisher={IEEE}
}

@article{monji2025robust,
  title={Robust-DefReg: a robust coarse to fine non-rigid point cloud registration method based on graph convolutional neural networks},
  author={Monji-Azad, Sara and Kinz, Marvin and Maennel, David and Scherl, Claudia and Hesser, Juergen},
  journal={Measurement Science and Technology},
  volume={36},
  number={1},
  pages={015426},
  year={2025},
  publisher={IOP Publishing}
}

@article{zhang2024point,
  title={Point cloud registration in laparoscopic liver surgery using keypoint correspondence registration network},
  author={Zhang, Yirui and Zou, Yanni and Liu, Peter X},
  journal={IEEE Transactions on Medical Imaging},
  year={2024},
  publisher={IEEE}
}

@inproceedings{jiang1992new,
  title={New approach to 3-D registration of multimodality medical images by surface matching},
  author={Jiang, Hongjian and Robb, Richard A and Tainter, Kerrie S Holton},
  booktitle={Visualization in Biomedical Computing'92},
  volume={1808},
  pages={196--213},
  year={1992},
  organization={SPIE}
}

@article{labrunie2022automatic,
  title={Automatic preoperative 3d model registration in laparoscopic liver resection},
  author={Labrunie, Mathieu and Ribeiro, Mathieu and Mourthadhoi, Farouk and Tilmant, Christophe and Le Roy, Bertrand and Buc, Emmanuel and Bartoli, Adrien},
  journal={International Journal of Computer Assisted Radiology and Surgery},
  volume={17},
  number={8},
  pages={1429--1436},
  year={2022},
  publisher={Springer}
}

@article{liu2023ranerf,
  title={RaNeRF: Neural 3-D reconstruction of space targets from ISAR image sequences},
  author={Liu, Afei and Zhang, Shuanghui and Zhang, Chi and Zhi, Shuaifeng and Li, Xiang},
  journal={IEEE Transactions on Geoscience and Remote Sensing},
  volume={61},
  pages={1--15},
  year={2023},
  publisher={IEEE}
}

\end{document}